\pdfoutput=1

\documentclass[11pt]{article}

\usepackage[]{ACL2023}

\usepackage{times}
\usepackage{latexsym}

\usepackage{amsmath}

\usepackage[skins]{tcolorbox}

\usepackage{algorithm2e}

\usepackage[T1]{fontenc}

\usepackage[utf8]{inputenc}

\usepackage{microtype}

\usepackage{inconsolata}

\usepackage{array}

\usepackage{graphicx}
\usepackage{arydshln}

\usepackage{amssymb}
\usepackage{pifont}
\usepackage{xcolor}
\usepackage{color,soul}

\newcommand{\cmark}{\ding{51}}%
\newcommand{\xmark}{\ding{53}}%

\definecolor{myblue}{RGB}{65, 100, 166}
\definecolor{mycyan}{RGB}{75, 186, 186}

\newtcolorbox{instructionframe}[2][]{%
  enhanced,colback=white,colframe=myblue,coltitle=white,boxrule=1.0pt,
  fonttitle=\mdseries,
  attach boxed title to top left={yshift=-0.5\baselineskip-0.4pt,xshift=2mm},
  boxed title style={tile,size=minimal,left=1.5mm,right=1.5mm,
    colback=myblue,before upper=\strut},
  title=#2,#1
}

\newtcolorbox{pseudocodeframe}[2][]{%
  enhanced,colback=white,colframe=mycyan,coltitle=white,boxrule=1.0pt,
  fonttitle=\mdseries,
  attach boxed title to top left={yshift=-0.5\baselineskip-0.4pt,xshift=2mm},
  boxed title style={tile,size=minimal,left=1.5mm,right=1.5mm,
    colback=mycyan,before upper=\strut},
  title=#2,#1
}

\newenvironment{enumerate*}
    {\begin{enumerate}%
      \setlength{\itemsep}{0pt}%
      \setlength{\parskip}{0pt}}%
    {\end{enumerate}}

\title{BioMistral: A Collection of Open-Source Pretrained Large Language Models for Medical Domains}

\author{Yanis Labrak{\normalfont $^*$\textsuperscript{1,2}} \hspace{0.5cm} Adrien Bazoge{\normalfont $^*$\textsuperscript{3,4}}\\ {\bf Emmanuel Morin}\textsuperscript{4} \hspace{0.5cm} {\bf Pierre-Antoine Gourraud}\textsuperscript{3} \hspace{0.5cm} {\bf Mickael Rouvier} {\normalfont \textsuperscript{1}}   \hspace{0.5cm} {\bf Richard Dufour} {\normalfont \textsuperscript{1,4}} \\ 
\textsuperscript{1}LIA, Avignon Université  \>  \textsuperscript{2}Zenidoc  \\ \textsuperscript{3}Nantes Université, CHU Nantes, Clinique des données, INSERM, CIC 1413, F-44000 Nantes, France \\ \textsuperscript{4}Nantes Université, École Centrale Nantes, CNRS, LS2N, UMR 6004, F-44000 Nantes, France   \\
\texttt{\{firstname.lastname\}@univ-avignon.fr} \\ \texttt{\{firstname.lastname\}@univ-nantes.fr}
}

\begin{document}
\maketitle
\def\thefootnote{*}\footnotetext{Equal contribution.}\def\thefootnote{\arabic{footnote}}
\begin{abstract}

Large Language Models (LLMs) have demonstrated remarkable versatility in recent years, offering potential applications across specialized domains such as healthcare and medicine. Despite the availability of various open-source LLMs tailored for health contexts, adapting general-purpose LLMs to the medical domain presents significant challenges.
In this paper, we introduce BioMistral, an open-source LLM tailored for the biomedical domain, utilizing Mistral as its foundation model and further pre-trained on PubMed Central. We conduct a comprehensive evaluation of BioMistral on a benchmark comprising 10 established medical question-answering (QA) tasks in English. We also explore lightweight models obtained through quantization and model merging approaches. Our results demonstrate BioMistral's superior performance compared to existing open-source medical models and its competitive edge against proprietary counterparts. Finally, to address the limited availability of data beyond English and to assess the multilingual generalization of medical LLMs, we automatically translated and evaluated this benchmark into 7 other languages. This marks the first large-scale multilingual evaluation of LLMs in the medical domain. Datasets, multilingual evaluation benchmarks, scripts, and all the models obtained during our experiments are freely released.



\end{abstract}

\section{Introduction}

In the rapidly evolving landscape of Natural Language Processing (NLP), generative Large Language Models (LLMs) like ChatGPT~\cite{openai2023chatgpt} and Vicuna~\cite{zheng2023judging} have revolutionized human-computer interactions, demonstrating remarkable versatility and advanced capabilities across various tasks and domains. These models exhibit human-like comprehension and reasoning, enabling them to tackle basic textual understanding as well as complex problem-solving tasks.
The emergence of open-source LLMs such as BLOOM~\cite{workshop2023bloom} and LLaMA~\cite{touvron2023llama1} underscores the transformative potential of these models, facilitating their innovative use in specialized domains including medicine~\cite{Dave2023-ld}.

However, integrating LLMs into healthcare and medicine presents unique challenges and opportunities~\cite{he2023survey,zhou2024survey}. While preliminary adoption has opened new avenues for innovation, concerns about data privacy risks associated with proprietary models like MedPaLM-2~\cite{singhal2023expertlevel} and GPT-4~\cite{nori2023capabilities} have arisen.
The community's interest in specialized LLMs for healthcare has led to initiatives like PMC-LLaMA~\cite{wu2023pmcllama} and MedAlpaca~\cite{han2023medalpaca}. However, the adoption of open-source medical models has been limited, primarily due to the lack of lightweight models allowing commercial use with performance comparable to larger or proprietary models. To address this gap, there is a need to develop specialized models based on open-source foundation ones like GPT-NeoX~\cite{black-etal-2022-gpt}, LLaMa 2~\cite{touvron2023llama}, or Mistral~\cite{jiang2023mistral}, and to optimize them for use on consumer-grade devices while maintaining performance.

In this work, we present BioMistral 7B, a specialized LLM tailored for the biomedical domain, derived from Mistral 7B Instruct v0.1~\cite{jiang2023mistral} and further pre-trained on PubMed Central.  Our contributions include: \vspace{-2mm}

\begin{enumerate*}
\item The construction of BioMistral 7B, the first open-source Mistral-based model for the biomedical domain, including the analysis of different evaluation strategies such as few-shot in-context learning and supervised fine-tuning.

\item An original study with the introduction of a benchmark of $10$ medical question-answering (QA) tasks in English, automatically translated into $7$ other languages, facilitating assessment against existing state-of-the-art open-source medical LLMs, and shedding light on its multilingual potential and robustness across diverse linguistic contexts.

\item A large in-depth quantitative analysis of the model's truthfulness and calibration in multilingual contexts. 

\item A rigorous evaluation of a suite of lightweight models derived through various efficient quantization approaches.

\item An exploration of novel model merging techniques between Mistral 7B Instruct and BioMistral 7B models, allowing leveraging the commonsense reasoning skills of specialized and general-purpose LLMs.

\end{enumerate*}\vspace{-2mm}

All datasets, the multilingual benchmark, pre-processing scripts, and models are accessible on HuggingFace\footnote{\href{https://huggingface.co/BioMistral}{https://huggingface.co/BioMistral}} and GitHub\footnote{\href{https://github.com/BioMistral/BioMistral}{https://github.com/BioMistral/BioMistral}} under Apache 2.0 license.

\section{Related Work}


Generative LLMs with a generalist purpose, such as GPT-4~\cite{openai2023gpt4} and Gemini~\cite{geminiteam2023gemini}, have demonstrated exceptional performance across various NLP tasks, including QA, text summarization, and language translation in zero- and few-shot scenarios. 
While these models firstly remain proprietary, the open-source movement has played a crucial role in democratizing LLMs, offering powerful capabilities to the broader community with Llama 2~\cite{touvron2023llama}, Vicuna~\cite{zheng2023judging}, Falcon~\cite{almazrouei2023falcon} and Mistral~\cite{jiang2023mistral} models. 



The adaptation of LLMs to specialized domains often involves encoding domain-specific knowledge to fully leverage the models' capabilities. This adaptation process has led to the development of several open-source models from pre-trained general domain models. Encoder-based models, such as BERT~\cite{devlin-etal-2019-bert}, and encoder-decoder, such as T5~\cite{2020t5}, have been adapted for the medical domain~\cite{10.1093/bioinformatics/btz682,10.1145/3458754,yuan-etal-2022-biobart,lu-etal-2022-clinicalt5} but faced challenges with QA tasks. More recently, multiple LLMs based on decoder-only architecture have been developed for the medical domain, such as BioGPT~\cite{Luo_2022} based on GPT-2~\cite{radford2019language}, ClinicalGPT~\cite{wang2023clinicalgpt}, based on BLOOM-7B~\cite{workshop2023bloom}, PMC-LLaMA~\cite{wu2023pmcllama}, MedAlpaca~\cite{han2023medalpaca} and Chat-Doctor~\cite{li2023chatdoctor}, based on LLaMA~\cite{touvron2023llama1}, and BioMedGPT-LM-7B~\cite{zhang2024biomedgpt} and  MediTron-7B~\cite{chen2023meditron70b}, adapted from Llama-2~\cite{touvron2023llama}. In contrast, proprietary medical LLMs like GPT-4 MedPrompt~\cite{nori2023generalist} and Med-PALM 2~\cite{singhal2023expertlevel} face usability issues similar to general-purpose models. Nonetheless, they can be used as benchmark "target goals" for the development of domain-specific open-source models.



\section{BioMistral}

In this section, we present the modules that facilitated the construction of BioMistral 7B. We first develop our training corpus (Section~\ref{pre_training_dataset}) used during further pre-training. We then present the model adaptation method (Section~\ref{model_adaptation}). Finally, we discuss the approaches for model merging (Section~\ref{sec:model_merging}) and expose the employed quantization strategies (Section~\ref{sec:model_quantization}).

\subsection{Pre-training Dataset}\label{pre_training_dataset}

For LLM adaptation to the medical domain, we selected the PMC Open Access Subset\footnote{\href{https://www.ncbi.nlm.nih.gov/pmc/tools/openftlist/}{https://www.ncbi.nlm.nih.gov/pmc/tools/openftlist/}} for its comprehensive and freely accessible collection of medical research papers. This choice is guided by the success demonstrated by PMC-LLaMA~\cite{wu2023pmcllama}, PubMedBERT~\cite{10.1145/3458754}, and SciFive~\cite{phan2021scifive}, which have showcased significant enhancements in language modeling for medical applications. Our focus lies on the Commercial Use Allowed subset, encompassing documents licensed under various Creative Commons licenses (CC0, CC BY, CC BY-SA, and CC BY-ND). This subset ensures the reusability of our model's outputs, even for commercial purposes.

In the preprocessing phase, we aim to optimize the dataset for training efficiency while considering hardware limitations. Our pre-training objective involves further pre-training Mistral on a subsample of this corpus, targeting 1.5 epochs within the 20-hour limit of Jean Zay HPC. This decision aligns with insights from the Zephyr model~\cite{tunstall2023zephyr}, which suggests that observing 1.5 times the corpus adequately enhances model performance, with marginal benefits beyond this threshold. We then meticulously selected 3 billion tokens from this pre-processed PubMed Central corpus, corresponding to roughly 1.47 million documents. The dataset comprises primarily English documents (98.75\% of the corpus), with the remaining portion encompassing $9$ languages, including Dutch, German, French, and others. Our strategy emphasizes a multilingual dataset approach by prioritizing non-English documents, supplemented with English texts, to ensure a diverse and representative training dataset to meet our 3 billion token target. The raw textual documents undergo pre-processing using the Mistral tokenizer, which includes tokenization and normalization processes.

\subsection{Model Adaptation}\label{model_adaptation}

\paragraph{Training details} We leverage Mistral 7B Instruct v0.1~\cite{jiang2023mistral} as the base model for adaptation due to its design tailored for incorporating instructions in prompts and its capacity for fine-tuning across diverse tasks using limited datasets. Pre-training settings for BioMistral 7B largely align with Mistral 7B Instruct v0.1.
For optimization, we employ the AdamW~\cite{loshchilov2018decoupled} optimizer alongside a cosine learning rate scheduler. Our model architecture inherits the standard transformer architecture from Mistral, including features such as Grouped-Query Attention~\cite{ainslie-etal-2023-gqa}, Sliding Window Attention~\cite{beltagy2020longformer} and Rolling Buffer Cache. We maintain an input context length of 2,048 tokens across all models, including the quantized versions (see Section~\ref{sec:model_quantization}), in conjunction with FlashAttention-2~\cite{dao2023flashattention2}.
For optimization, we set the learning rate to $2 \times 10^{-5}$ with no warmup, a weight decay of $0.01$, a gradient accumulation of $2$, and a batch size of $16$ on the Jean-Zay HPC with 32 NVIDIA A100 80GB GPUs. This configuration allows for a total batch size of 1,024. Due to the model and the AdamW optimizer's inability to fit on a single GPU with BF16 precision, we employ the Fully Sharded Data Parallel distributed learning framework~\cite{zhao2023pytorch}. 

\paragraph{Improving batching}\label{improving_batching} To enhance pre-training efficiency, we introduce a post-tokenization grouping method. This method aggregates variable-sized sequences marked by an end-of-sequence token (</s>) to fill the model's 2,048-token sequences without padding. This reduces the sequence count by 87.88\%, subsequently accelerating epoch times. Refer to Appendix~\ref{sec:pseudocode-groupped} for pseudo-code detailing the grouping method.

\begin{table*}[]
\tiny
\setlength\extrarowheight{4pt}
\centering
\resizebox{\textwidth}{!}{%
\begin{tabular}{cccccccccc}
\hline
 &
  \multicolumn{6}{c}{\textbf{MMLU}} &
   &
   &
   \\ \cline{2-7}
 &
  \textbf{Clinical KG} &
  \textbf{Medical Genetics} &
  \textbf{Anatomy} &
  \textbf{Pro Medicine} &
  \textbf{College Biology} &
  \textbf{College Medicine} &
  \textbf{MedQA} &
  \textbf{PubMedQA} &
  \textbf{MedMCQA} \\ \hline
Answer options &
  A / B / C / D &
  A / B / C / D &
  A / B / C / D &
  A / B / C / D &
  A / B / C / D &
  A / B / C / D &
  A / B / C / D / (E) &
  Yes / No / Maybe &
  A / B / C / D \\
Train / Valid. / Test &
  0 / 0 / 265 &
  0 / 0 / 100 &
  0 / 0 / 135 &
  0 / 0 / 272 &
  0 / 0 / 144 &
  0 / 0 / 173 &
  10178 / 1272 / 1273 &
  211269 / 500 / 500 &
  146257 / 36565 / 4183 \\
Words / Questions &
  11.09 &
  12.34 &
  13.65 &
  105.46 &
  22.40 &
  48.84 &
  118.16 &
  13.08 &
  14.05 \\
Context &
  \xmark &
  \xmark &
  \xmark &
  \xmark &
  \xmark &
  \xmark &
  \xmark &
  \cmark &
  \xmark \\
\hline
\end{tabular}%
}
\caption{Description of the benchmarked question-answering tasks, including the number of train, validation, and test questions, and answer options for each task. Only PubMedQA incorporates context information within the prompt (see Appendix~\ref{sec:Prompting}). The reference to "Clinical KG" denotes "Clinical Knowledge".}

\label{tab:Datasets}
\end{table*}

\subsection{Model Merging}\label{sec:model_merging}

Pre-trained models may lose effectiveness when applied beyond their specific domains~\cite{labrak-etal-2023-drbert}. Traditionally, separate models were used for each application~\cite{guo-etal-2018-multi}, increasing complexity and costs. Recent studies suggest merging pre-trained models to enhance performance and out-of-domain generalization~\cite{NEURIPS2021_bcb41ccd, arpit2022ensemble, pmlr-v162-wortsman22a, jin2023dataless, ilharco2023editing}.
Merging involves combining multiple model parameters without additional training. Methods include averaging model weights or considering permutation invariance~\cite{ilharco2022patching, choshen2022fusing, NEURIPS2020_fb269786, ainsworth2023git}.

Among these methods, we can cite TIES~\cite{yadav2023tiesmerging}, DARE~\cite{yu2024language}, and SLERP~\cite{10.1145/325334.325242}. SLERP merges two models using Spherical Linear Interpolation to allow a smoother transition between model parameters while preventing the significant information loss often encountered with direct averaging of model weights. TIES merges models by creating "task vectors" from each model, isolating unique contributions by subtracting an ancestor base model (e.g., Mistral 7B Instruct). These vectors are then averaged with the base model.
Its key improvement over previous methods relies on reducing model interference using sparse vectors and a sign consensus method.
DARE enhances TIES by reducing delta parameter redundancy, mainly setting them to zero through random pruning and rescaling while maintaining or improving original model performance.

Exploring model merging in the biomedical domain is particularly interesting since merging a general domain model with a domain-specific one could enhance specialized model adaptability and accuracy across a broader range of applications.
The objective of this application in the medical domain is not only to improve general-domain capabilities but also to explore the possibility of emergent reasoning and surpassing the performance of baseline models used for merging.

\subsection{Quantization}\label{sec:model_quantization}

Quantization techniques are pivotal in democratizing LLMs as they enable the execution of LLMs on smaller devices by minimizing memory requirements. In our study, we investigate two core techniques: Activation-aware Weight Quantization (AWQ) and BitsandBytes (BnB). AWQ~\cite{lin2023awq} capitalizes on the insight that weights vary in importance, allowing us to skip quantizing critical weights to mitigate performance degradation. Conversely, BnB quantization assigns a fixed precision of 4 or 8 bits to the entire model.



\section{Evaluation Protocol}
\label{evaluation_tasks}

To assess the performance of BioMistral 7B models, we first describe our benchmark of English medical reasoning tasks (Section~\ref{downstream_tasks}) and their multilingual translation (Section~\ref{sec:multilingualdataset}), before presenting the instruction prompting (Section~\ref{sec:PromptingInstruction}) and the supervised fine-tuning strategy (Section~\ref{sec:SFT}) employed for the models' evaluation.

\begin{table*}[]
\setlength\extrarowheight{4pt}
\centering
\resizebox{\textwidth}{!}{%
\begin{tabular}{llllllllllll}
\hline
  &
  \multicolumn{6}{c}{\textbf{MMLU}} &
    &
    &
    &
    \\ \cline{2-7}
  &
  \textbf{Clinical KG} &
  \textbf{Medical Genetics} &
  \textbf{Anatomy} &
  \textbf{Pro Medicine} &
  \textbf{College Biology} &
  \textbf{College Medicine} &
  \textbf{MedQA} &
  \textbf{MedQA 5 opts} &
  \textbf{PubMedQA} &
  \textbf{MedMCQA} & \textbf{Avg.} \\ \hline
  
\textbf{BioMistral 7B} &
  60.9 \scalebox{1.0}{\tiny {±1.5}} &
  61.7 \scalebox{1.0}{\tiny {±2.1}} &
  \underline{49.6} \scalebox{1.0}{\tiny {±1.2}} &
  55.1 \scalebox{1.0}{\tiny {±1.3}} &
  56.9 \scalebox{1.0}{\tiny {±1.0}} &
  55.5 \scalebox{1.0}{\tiny {±1.7}} &
  44.4 \scalebox{1.0}{\tiny {±0.2}} &
  37.4 \scalebox{1.0}{\tiny {±0.4}} &
  37.6 \scalebox{1.0}{\tiny {±1.5}} &
  43.9 \scalebox{1.0}{\tiny {±0.3}} & 50.3 \\
  
\textbf{Mistral 7B Instruct} &
57.0 \scalebox{1.0}{\tiny {±0.8}} &
56.7 \scalebox{1.0}{\tiny {±0.5}} &
46.9 \scalebox{1.0}{\tiny {±0.3}} &
51.0 \scalebox{1.0}{\tiny {±1.1}} &
58.6 \scalebox{1.0}{\tiny {±0.9}} &
50.1 \scalebox{1.0}{\tiny {±1.0}} &
42.3 \scalebox{1.0}{\tiny {±0.3}} &
34.5 \scalebox{1.0}{\tiny {±0.5}} &
\textbf{72.2} \scalebox{1.0}{\tiny {±0.5}} &
42.8 \scalebox{1.0}{\tiny {±0.5}} & 51.2 \\

\hdashline

\textbf{BioMistral 7B Ensemble} & \underline{62.8} \scalebox{1.0}{\tiny {±0.5}} & \underline{62.7} \scalebox{1.0}{\tiny {±1.7}} & 46.9 \scalebox{1.0}{\tiny {±0.3}} & 57.0 \scalebox{1.0}{\tiny {±0.6}} & 60.6 \scalebox{1.0}{\tiny {±0.9}} & 56.3 \scalebox{1.0}{\tiny {±0.3}} & 44.7 \scalebox{1.0}{\tiny {±0.4}} & 37.1 \scalebox{1.0}{\tiny {±0.6}} & 68.0 \scalebox{1.0}{\tiny {±0.4}} & 44.8 \scalebox{1.0}{\tiny {±0.3}} & 54.1 \\


\textbf{BioMistral 7B DARE} & 61.3 \scalebox{1.0}{\tiny {±0.4}} & 61.0 \scalebox{1.0}{\tiny {±2.8}} & \textbf{49.9} \scalebox{1.0}{\tiny {±0.9}} & 55.3 \scalebox{1.0}{\tiny {±0.7}} & \textbf{64.4} \scalebox{1.0}{\tiny {±0.9}} & 53.9 \scalebox{1.0}{\tiny {±1.4}} & \textbf{47.0} \scalebox{1.0}{\tiny {±0.5}} & \underline{38.8} \scalebox{1.0}{\tiny {±0.7}} & \underline{70.0} \scalebox{1.0}{\tiny {±0.7}} & \underline{44.9} \scalebox{1.0}{\tiny {±0.2}} & \underline{54.6} \\

\textbf{BioMistral 7B TIES} & 62.3 \scalebox{1.0}{\tiny {±0.5}} & 61.3 \scalebox{1.0}{\tiny {±1.9}} & 48.1 \scalebox{1.0}{\tiny {±2.2}} & 55.8 \scalebox{1.0}{\tiny {±0.8}} & 57.2 \scalebox{1.0}{\tiny {±0.7}} & \underline{56.5} \scalebox{1.0}{\tiny {±1.5}} & 44.0 \scalebox{1.0}{\tiny {±0.4}} & 37.7 \scalebox{1.0}{\tiny {±0.4}} & 44.3 \scalebox{1.0}{\tiny {±0.8}} & 44.0 \scalebox{1.0}{\tiny {±0.3}} & 51.1 \\

\textbf{BioMistral 7B SLERP} & \textbf{63.1} \scalebox{1.0}{\tiny {±1.6}} & \textbf{63.3} \scalebox{1.0}{\tiny {±0.9}} & \textbf{49.9} \scalebox{1.0}{\tiny {±1.9}} & \underline{57.4} \scalebox{1.0}{\tiny {±0.3}} & \underline{63.4} \scalebox{1.0}{\tiny {±0.9}} & \textbf{57.8} \scalebox{1.0}{\tiny {±0.9}} & \underline{46.6} \scalebox{1.0}{\tiny {±0.2}} & \textbf{38.9} \scalebox{1.0}{\tiny {±0.4}} & 68.1 \scalebox{1.0}{\tiny {±1.4}} & \textbf{45.7} \scalebox{1.0}{\tiny {±0.7}} & \textbf{55.4} \\

\hdashline

\textbf{MedAlpaca 7B} & 49.1 \scalebox{1.0}{\tiny {±1.3}} & 49.0 \scalebox{1.0}{\tiny {±5.7}} & 48.4 \scalebox{1.0}{\tiny {±1.9}} & \textbf{63.8} \scalebox{1.0}{\tiny {±0.8}} & 47.2 \scalebox{1.0}{\tiny {±0.6}} & 43.5 \scalebox{1.0}{\tiny {±1.8}} & 35.4 \scalebox{1.0}{\tiny {±0.3}} & 30.4 \scalebox{1.0}{\tiny {±0.6}} & 56.0 \scalebox{1.0}{\tiny {±0.9}} & 31.2 \scalebox{1.0}{\tiny {±0.2}} & 45.4 \\

\textbf{PMC-LLaMA 7B} & 25.3 \scalebox{1.0}{\tiny {±1.5}} & 26.0 \scalebox{1.0}{\tiny {±3.7}} & 31.9 \scalebox{1.0}{\tiny {±1.8}} & 16.9 \scalebox{1.0}{\tiny {±0.5}} & 28.0 \scalebox{1.0}{\tiny {±2.4}} & 24.9 \scalebox{1.0}{\tiny {±1.2}} & 27.6 \scalebox{1.0}{\tiny {±0.8}} & 21.1 \scalebox{1.0}{\tiny {±0.8}} & 53.3 \scalebox{1.0}{\tiny {±0.6}} & 23.5 \scalebox{1.0}{\tiny {±0.3}} & 27.8 \\

\textbf{MediTron-7B} &
  37.9 \scalebox{1.0}{\tiny {±1.5}} &
  47.0 \scalebox{1.0}{\tiny {±3.7}} &
  39.3 \scalebox{1.0}{\tiny {±1.6}} &
  34.2 \scalebox{1.0}{\tiny {±1.0}} &
  42.6 \scalebox{1.0}{\tiny {±1.4}} &
  30.4 \scalebox{1.0}{\tiny {±0.7}} &
  34.8 \scalebox{1.0}{\tiny {±0.6}} &
  26.3 \scalebox{1.0}{\tiny {±0.5}} &
  55.9 \scalebox{1.0}{\tiny {±1.0}} &
  33.6 \scalebox{1.0}{\tiny {±0.2}} & 38.2 \\

\textbf{BioMedGPT-LM-7B} & 50.1 \scalebox{1.0}{\tiny {±1.0}} & 52.0 \scalebox{1.0}{\tiny {±0.8}} & 46.2 \scalebox{1.0}{\tiny {±1.8}} & 47.3 \scalebox{1.0}{\tiny {±1.7}} & 47.9 \scalebox{1.0}{\tiny {±2.5}} & 45.5 \scalebox{1.0}{\tiny {±0.7}} & 39.3 \scalebox{1.0}{\tiny {±1.2}} & 34.9 \scalebox{1.0}{\tiny {±0.4}} & 58.6 \scalebox{1.0}{\tiny {±0.3}} & 34.9 \scalebox{1.0}{\tiny {±0.5}} & 45.7 \\

\hdashline
  
\textbf{GPT-3.5 Turbo 1106} &
  74.71 \scalebox{1.0}{\tiny {±0.3}} &
  74.00 \scalebox{1.0}{\tiny {±2.2}} &
  65.92 \scalebox{1.0}{\tiny {±0.6}} &
  72.79 \scalebox{1.0}{\tiny {±1.6}} &
  72.91 \scalebox{1.0}{\tiny {±1.7}} &
  64.73 \scalebox{1.0}{\tiny {±2.9}} &
  
  57.71 \scalebox{1.0}{\tiny {±0.3}} &
  50.82 \scalebox{1.0}{\tiny {±0.7}} &
  72.66 \scalebox{1.0}{\tiny {±1.0}} &
  53.79 \scalebox{1.0}{\tiny {±0.2}} & 66.0 \\
  \hline
\end{tabular}%
}
\caption{Performance of 3-shot in-context learning. The scores represent accuracy ($\uparrow$) and are averaged across 3 random seeds. BioMistral 7B Ensemble, DARE, TIES, and SLERP are model merging strategies that combine BioMistral 7B and Mistral 7B Instruct. Best model in bold, and second-best underlined.}
\label{tab:few-shot}
\end{table*}

\subsection{Downstream Tasks}\label{downstream_tasks}

To evaluate the performance of the BioMistral 7B model, we selected 10 QA tasks in English from 4 prominent medical corpora covering various specialties, including genetics, anatomy, and clinical cases. These datasets encapsulate real-world scenarios encountered by medical professionals, medical school entrance examination formats, and comprehension tests based on PubMed content. The datasets' characteristics are provided in Table~\ref{tab:Datasets}.

\paragraph{MMLU}~\cite{hendrycks2021measuring} comprises exam questions spanning 57 subjects, including philosophy, management, and medical domains. We focused on the 6 subjects relevant to medical and clinical knowledge, and already used to evaluate MedPaLM-2~\cite{singhal2023expertlevel}: college biology, college medicine, anatomy, professional medicine, medical genetics, and clinical knowledge. These subjects were amalgamated to form a consolidated medical-related benchmark, featuring 1,089 questions. As MMLU lacks training data, we fine-tuned our models on MedQA and evaluated their generalization performance on MMLU.

\paragraph{MedQA}~\cite{jin2020disease} presents questions in the format of the US Medical License Exam (USMLE) and encompasses diverse medical knowledge, including patient profiles, disease symptoms, and drug dosage requirements. The training set comprises 10,178 samples, while the test set contains 1,273 questions. MedQA provides two configurations: four-choice (MedQA) and five-choice (MedQA 5-options) question formats.

\paragraph{MedMCQA}~\cite{pmlr-v174-pal22a} consists of over 193k MCQA with 4 options each, extracted from Indian medical entrance examinations (AIIMS/NEET). It covers 2,400 healthcare topics across 21 medical subjects. The training set comprises 183k samples and the validation set includes 4,183 questions. Due to the unavailability of answer keys for the test set of 6,150 questions, we adopt a similar approach to~\citet{wu2023pmcllama}, using the validation set for evaluation. For hyperparameter tuning, we perform a random split of the training set into new train/validation splits with 146k and 37k samples each.

\paragraph{PubMedQA}~\cite{jin-etal-2019-pubmedqa} comprises 211k artificially generated multiple-choice question samples and 1,000 samples labeled by experts. In our evaluation, we adhere to the {\it reasoning-required} setting, in which the model has to predict {\it yes}, {\it no}, or {\it maybe} for a given PubMed abstract used as context and a corresponding question. Fine-tuning is performed using 211k artificially labeled samples and performance is accessed on 500 expert-labeled samples during validation and 500 during test as split in BigBio~\cite{NEURIPS2022_a583d219} and following~\citet{chen2023meditron70b,Singhal2023} protocol.

\subsection{Multilingual Evaluation}
\label{sec:multilingualdataset}

While the biomedical language models have been extensively evaluated in languages such as English~\cite{10.1093/bioinformatics/btz682, chen2023meditron70b}, Chinese~\cite{10.1007/978-3-030-85896-4_20, yang2023zhongjing}, French~\cite{touchent-etal-2023-camembert, labrak-etal-2023-drbert} or Spanish~\cite{carrino-etal-2022-pretrained}, their performance in languages beyond their own remains relatively understudied. This limited multilingual evaluation can be attributed to the scarcity of biomedical tasks available in languages other than English. To address this gap, we conducted a multilingual evaluation using GPT-3.5 Turbo (version 1106) automatic translation via the OpenAI API. We translated our benchmark into 7 languages: Spanish, German, Portuguese, Russian, French, Arabic, and Chinese. Despite the challenges posed by automatic translation, these tools have shown remarkable improvement in recent years~\cite{neves-etal-2023-findings}, enabling cost-effective multilingual evaluation. The methodology for multilingual evaluation and the prompt template are the same as those used in the 3-shot scenario for English. The only differences lie in the translation of the questions, options, and context, while the examples used for few-shot learning remain unchanged.

\subsection{Instruction Prompting}
\label{sec:PromptingInstruction}

All of our instructions adhere to the guidelines outlined for GPT-4's medical evaluation, as detailed in~\citet{nori2023capabilities}. Each task is presented as an MCQA, with answer options associated with letters (A to D or A to E). For a comprehensive list of the instruction prompts, please refer to Appendix~\ref{sec:Prompting}. During inference, the model predicts the next token based on the input prompt, generating probabilities for each token in the vocabulary. To ensure relevance, the vocabulary is filtered to include only tokens (here, choice letters) corresponding to the expected answer options. This approach prevents the model from generating irrelevant tokens or hallucinations~\cite{liang2023holistic,beeching2023open,chen2023meditron70b}.

\begin{table*}[]
\centering
\setlength\extrarowheight{4pt}
\resizebox{\textwidth}{!}{%
\begin{tabular}{llllllllllll}
\hline
&
\multicolumn{7}{c}{\textbf{MMLU}} &
&
&
&
\\ \cline{2-7}
&
\textbf{Clinical KG} &
\textbf{Medical Genetics} &
\textbf{Anatomy} &
\textbf{Pro Medicine} &
\textbf{College Biology} &
\textbf{College Medicine} &
\textbf{MedQA} &
\textbf{MedQA 5 opts} &
\textbf{PubMedQA} &
\textbf{MedMCQA} & \textbf{Avg.} \\ \hline

\textbf{BioMistral 7B} &
59.9 \scalebox{1.0}{\tiny {±1.2}} & 64.0 \scalebox{1.0}{\tiny {±1.6}} & {56.5} \scalebox{1.0}{\tiny {±1.8}} & 60.4 \scalebox{1.0}{\tiny {±0.5}} & 59.0 \scalebox{1.0}{\tiny {±1.5}} & 54.7 \scalebox{1.0}{\tiny {±1.0}}  &

50.6 \scalebox{1.0}{\tiny {±0.3}} &
42.8 \scalebox{1.0}{\tiny {±0.3}} &
77.5 \scalebox{1.0}{\tiny {±0.1}} &
48.1 \scalebox{1.0}{\tiny {±0.2}} & 57.3 \\

\textbf{Mistral 7B Instruct} &
\textbf{62.9} \scalebox{1.0}{\tiny {±0.2}} & 57.0 \scalebox{1.0}{\tiny {±0.8}} & 55.6 \scalebox{1.0}{\tiny {±1.0}} & 59.4 \scalebox{1.0}{\tiny {±0.6}} & 62.5 \scalebox{1.0}{\tiny {±1.0}} & \underline{57.2} \scalebox{1.0}{\tiny {±2.1}} &

42.0 \scalebox{1.0}{\tiny {±0.2}} &
40.9 \scalebox{1.0}{\tiny {±0.4}} &
75.7 \scalebox{1.0}{\tiny {±0.4}} &
46.1 \scalebox{1.0}{\tiny {±0.1}} & 55.9 \\

\hdashline

\textbf{BioMistral 7B Ensemble} & \underline{62.8} \scalebox{1.0}{\tiny {±0.5}} & 62.7 \scalebox{1.0}{\tiny {±0.5}} & \underline{57.5} \scalebox{1.0}{\tiny {±0.3}} & \textbf{63.5} \scalebox{1.0}{\tiny {±0.8}} & 64.3 \scalebox{1.0}{\tiny {±1.6}} & 55.7 \scalebox{1.0}{\tiny {±1.5}} & 50.6 \scalebox{1.0}{\tiny {±0.3}} & 43.6 \scalebox{1.0}{\tiny {±0.5}} & 77.5 \scalebox{1.0}{\tiny {±0.2}} & \textbf{48.8} \scalebox{1.0}{\tiny {±0.0}} & 58.7 \\

\textbf{BioMistral 7B DARE} &
62.3 \scalebox{1.0}{\tiny {±1.3}} & \textbf{67.0} \scalebox{1.0}{\tiny {±1.6}} & 55.8 \scalebox{1.0}{\tiny {±0.9}} & {61.4} \scalebox{1.0}{\tiny {±0.3}} & \textbf{66.9} \scalebox{1.0}{\tiny {±2.3}} & \textbf{58.0} \scalebox{1.0}{\tiny {±0.5}} &
\textbf{51.1} \scalebox{1.0}{\tiny {±0.3}} &
\textbf{45.2} \scalebox{1.0}{\tiny {±0.3}} &
\underline{77.7} \scalebox{1.0}{\tiny {±0.1}} &
\underline{48.7} \scalebox{1.0}{\tiny {±0.1}} & \textbf{59.4} \\ 

\textbf{BioMistral 7B TIES} &
60.1 \scalebox{1.0}{\tiny {±0.9}} & \underline{65.0} \scalebox{1.0}{\tiny {±2.4}} & \textbf{58.5} \scalebox{1.0}{\tiny {±1.0}} & 60.5 \scalebox{1.0}{\tiny {±1.1}} & 60.4 \scalebox{1.0}{\tiny {±1.5}} & 56.5 \scalebox{1.0}{\tiny {±1.9}} &
49.5 \scalebox{1.0}{\tiny {±0.1}} &
43.2 \scalebox{1.0}{\tiny {±0.1}} &
77.5 \scalebox{1.0}{\tiny {±0.2}} &
48.1 \scalebox{1.0}{\tiny {±0.1}} & 57.9 \\ 

\textbf{BioMistral 7B SLERP} &
{62.5} \scalebox{1.0}{\tiny {±0.6}} & 64.7 \scalebox{1.0}{\tiny {±1.7}} & 55.8 \scalebox{1.0}{\tiny {±0.3}} & \underline{62.7} \scalebox{1.0}{\tiny {±0.3}} & \underline{64.8} \scalebox{1.0}{\tiny {±0.9}} & 56.3 \scalebox{1.0}{\tiny {±1.0}} &
\underline{50.8} \scalebox{1.0}{\tiny {±0.6}} &
\underline{44.3} \scalebox{1.0}{\tiny {±0.4}} &
\textbf{77.8} \scalebox{1.0}{\tiny {±0.0}} &
{48.6} \scalebox{1.0}{\tiny {±0.1}} & \underline{58.8} \\ 

\hdashline

\textbf{MedAlpaca 7B} &
53.1 \scalebox{1.0}{\tiny {±0.9}} & 58.0 \scalebox{1.0}{\tiny {±2.2}} & 54.1 \scalebox{1.0}{\tiny {±1.6}} & 58.8 \scalebox{1.0}{\tiny {±0.3}} & 58.1 \scalebox{1.0}{\tiny {±1.3}} & 48.6 \scalebox{1.0}{\tiny {±0.5}} &

40.1 \scalebox{1.0}{\tiny {±0.4}} &
33.7 \scalebox{1.0}{\tiny {±0.7}} &
73.6 \scalebox{1.0}{\tiny {±0.3}} &
37.0 \scalebox{1.0}{\tiny {±0.3}} & 51.5 \\

\textbf{PMC-LLaMA 7B} &
24.5 \scalebox{1.0}{\tiny {±1.7}} & 27.7 \scalebox{1.0}{\tiny {±1.7}} & 35.3 \scalebox{1.0}{\tiny {±0.7}} & 17.4 \scalebox{1.0}{\tiny {±1.7}} & 30.3 \scalebox{1.0}{\tiny {±0.9}} & 23.3 \scalebox{1.0}{\tiny {±1.7}} &

25.5 \scalebox{1.0}{\tiny {±0.9}} &
20.2 \scalebox{1.0}{\tiny {±0.1}} &
72.9 \scalebox{1.0}{\tiny {±1.2}} &
26.6 \scalebox{1.0}{\tiny {±0.1}} & 30.4 \\

\textbf{MediTron-7B} &
 41.6 \scalebox{1.0}{\tiny {±1.2}} & 50.3 \scalebox{1.0}{\tiny {±2.1}} & 46.4 \scalebox{1.0}{\tiny {±0.9}} & 27.9 \scalebox{1.0}{\tiny {±0.3}} & 44.4 \scalebox{1.0}{\tiny {±2.6}} & 30.8 \scalebox{1.0}{\tiny {±0.7}}  &

41.6 \scalebox{1.0}{\tiny {±0.5}} &
28.1 \scalebox{1.0}{\tiny {±0.5}} &
74.9 \scalebox{1.0}{\tiny {±0.1}} &
41.3 \scalebox{1.0}{\tiny {±0.2}} & 42.7 \\

\textbf{BioMedGPT-LM-7B} &
51.4 \scalebox{1.0}{\tiny {±0.4}} & 52.0 \scalebox{1.0}{\tiny {±1.4}} & 49.4 \scalebox{1.0}{\tiny {±2.7}} & 53.3 \scalebox{1.0}{\tiny {±0.6}} & 50.7 \scalebox{1.0}{\tiny {±0.0}} & 49.1 \scalebox{1.0}{\tiny {±0.8}} &
42.5 \scalebox{1.0}{\tiny {±0.3}} &
33.9 \scalebox{1.0}{\tiny {±0.5}} &
76.8 \scalebox{1.0}{\tiny {±0.3}} &
37.6 \scalebox{1.0}{\tiny {±0.4}} & 49.7 \\

\hdashline
  
\textbf{GPT-3.5 Turbo 1106*} &
  74.71 \scalebox{1.0}{\tiny {±0.3}} &
  74.00 \scalebox{1.0}{\tiny {±2.2}} &
  65.92 \scalebox{1.0}{\tiny {±0.6}} &
  72.79 \scalebox{1.0}{\tiny {±1.6}} &
  72.91 \scalebox{1.0}{\tiny {±1.7}} &
  64.73 \scalebox{1.0}{\tiny {±2.9}} &
  
  57.71 \scalebox{1.0}{\tiny {±0.3}} &
  50.82 \scalebox{1.0}{\tiny {±0.7}} &
  72.66 \scalebox{1.0}{\tiny {±1.0}} &
  53.79 \scalebox{1.0}{\tiny {±0.2}} & 66.0 \\

\hline

\end{tabular}%
}
\caption{Supervised Fine-Tuning (SFT) performance of BioMistral 7B models compared to baselines, measured by accuracy ($\uparrow$) and averaged across 3 random seeds of 3-shot. DARE, TIES, and SLERP are model merging strategies that combine BioMistral 7B and Mistral 7B Instruct. Best model in bold, and second-best underlined. *GPT-3.5 Turbo performances are reported from the few-shot results in Table~\ref{tab:few-shot}.}
\label{tab:sft}
\end{table*}

\subsection{Supervised Fine-Tuning (SFT)}
\label{sec:SFT}

Supervised Fine-Tuning (SFT) is a crucial step involving fine-tuning the model on annotated data to adapt it to specific tasks. To optimize BioMistral's performance beyond what is achievable with few-shot learning, we conducted SFT on both BioMistral 7B models and the baseline open-source models, using the training sets specified in Table~\ref{tab:Datasets}. However, traditional SFT methods can be resource-intensive. To address this challenge, we adopted the QLoRa fine-tuning method~\cite{dettmers2023qlora} and an 8-bit quantization technique~\cite{dettmers2022gptint} as more cost-effective alternatives. Additionally, we implemented the improved batching method discussed in Section~\ref{improving_batching} to reduce fine-tuning time. For detailed hyperparameters used during SFT, please refer to Appendix~\ref{sec:SFT-Hyperparameters}.




\section{Results and Discussions}

In this section, we report, analyze, and discuss the performance of BioMistral 7B models across various dimensions. We begin by examining its performance in a few-shot learning scenario (Section~\ref{few_shot}), followed by an evaluation of the fine-tuning performances (Section~\ref{results_sft}) of BioMistral 7B compared to several baseline models. The effectiveness of BioMistral 7B model merging strategies is then reported (Section~\ref{results_model_merging}) before exploring its generalization capabilities across several languages (Section~\ref{results_multilingual}). Additionally, we analyze the performance of BioMistral quantized versions in a few-shot scenario (Section~\ref{results_quantization}). Finally, we delve into its reliability by examining its calibration (Section~\ref{results_calibration}) and truthfulness (Section~\ref{results_truthfulness}).

\subsection{Few-shot Learning}\label{few_shot}

The few-shot learning evaluation involved applying 3-shot in-context learning based on 3 different sets of randomly selected samples from each dataset's training set. We limited our samples to 3 due to the model's 2,048-token context window size. None of the models were fine-tuned on the datasets. 

In Table~\ref{tab:few-shot}, we observe that BioMistral 7B outperforms Mistral 7B Instruct on 8 of the 10 tasks, demonstrating the effectiveness of domain adaptation~\cite{chen2023meditron70b, 10.1093/bioinformatics/btz682}. Additionally, BioMistral 7B surpasses all other open-source biomedical baselines on all tasks in this 3-shot scenario. 
The observed performances may vary depending on the dataset. For example, on MedQA 4 and 5 options, BioMistral 7B shows a 9.6\% and 11.1\% increase over MediTron-7B and a 9.0\% and 7.0\% increase over MedAlpaca 7B, respectively. On MMLU, BioMistral 7B improves performance over previous biomedical LLMs at the 7B scale, with an overall average gain of 6.45\% over MedAlpaca 7B, 18.05\% over MediTron-7B, and 31.12\% over PMC-LLaMA 7B. Similarly, on MedMCQA, BioMistral 7B shows a 10.3\% increase over MediTron-7B, 12.7\% over MedAlpaca 7B, and 20.4\% over PMC-LLaMA 7B.
However, in the PubMedQA evaluation, BioMistral's performance experienced a decline, showing at least a 15.7\% lower accuracy compared to other models, likely due to hallucinations caused by imbalanced classes.
Overall, GPT-3.5 Turbo remains the best model in this 3-shot scenario.





\begin{table*}[]
\centering
\setlength\extrarowheight{4pt}
\resizebox{\textwidth}{!}{%
\begin{tabular}{llllllllllll}
\hline
&
\multicolumn{6}{c}{\textbf{MMLU}} &
&
&
&
\\ \cline{2-7}
&
\textbf{Clinical KG} &
\textbf{Medical Genetics} &
\textbf{Anatomy} &
\textbf{Pro Medicine} &
\textbf{College Biology} &
\textbf{College Medicine} &
\textbf{MedQA} &
\textbf{MedQA 5 opts} &
\textbf{PubMedQA} &
\textbf{MedMCQA} & \textbf{Avg.} \\ \hline

\textbf{BioMistral 7B*} &
60.9 \scalebox{1.0}{\tiny {±1.5}} &
61.7 \scalebox{1.0}{\tiny {±2.1}} &
49.6 \scalebox{1.0}{\tiny {±1.2}} &
55.1 \scalebox{1.0}{\tiny {±1.3}} &
56.9 \scalebox{1.0}{\tiny {±1.0}} &
55.5 \scalebox{1.0}{\tiny {±1.7}} &
44.4 \scalebox{1.0}{\tiny {±0.2}} &
37.4 \scalebox{1.0}{\tiny {±0.4}} &
37.6 \scalebox{1.0}{\tiny {±1.5}} &
43.9 \scalebox{1.0}{\tiny {±0.3}} & 50.3 \\

\hdashline

\textbf{AWQ 4bit + GEMV} & 59.5 \scalebox{1.0}{\tiny {±1.2}} & 61.3 \scalebox{1.0}{\tiny {±1.7}} & 50.6 \scalebox{1.0}{\tiny {±2.5}} & 53.9 \scalebox{1.0}{\tiny {±0.7}} & 56.2 \scalebox{1.0}{\tiny {±1.5}} & 52.6 \scalebox{1.0}{\tiny {±1.7}} & 43.2 \scalebox{1.0}{\tiny {±0.8}} & 36.8 \scalebox{1.0}{\tiny {±0.5}} & 61.7 \scalebox{1.0}{\tiny {±0.9}} & 41.8 \scalebox{1.0}{\tiny {±0.2}} & 51.8 \scalebox{1.0}{\tiny {+1.5}} \\

\textbf{AWQ 4bit + GEMM} & 59.5 \scalebox{1.0}{\tiny {±1.2}} & 61.3 \scalebox{1.0}{\tiny {±1.2}} & 50.6 \scalebox{1.0}{\tiny {±2.5}} & 53.6 \scalebox{1.0}{\tiny {±0.8}} & 56.2 \scalebox{1.0}{\tiny {±1.5}} & 52.4 \scalebox{1.0}{\tiny {±1.5}} & 43.2 \scalebox{1.0}{\tiny {±0.8}} & 37.0 \scalebox{1.0}{\tiny {±0.5}} & 61.4 \scalebox{1.0}{\tiny {±0.9}} & 41.8 \scalebox{1.0}{\tiny {±0.2}} & 51.7 \scalebox{1.0}{\tiny {+1.4}} \\

\hdashline

\textbf{BnB 4bit} & 57.6 \scalebox{1.0}{\tiny {±1.1}} & 58.7 \scalebox{1.0}{\tiny {±0.9}} & 47.2 \scalebox{1.0}{\tiny {±0.9}} & 52.9 \scalebox{1.0}{\tiny {±1.3}} & 53.7 \scalebox{1.0}{\tiny {±0.9}} & 54.3 \scalebox{1.0}{\tiny {±1.2}} & 43.1 \scalebox{1.0}{\tiny {±0.2}} & 36.8 \scalebox{1.0}{\tiny {±0.9}} & 22.4 \scalebox{1.0}{\tiny {±0.4}} & 42.0 \scalebox{1.0}{\tiny {±0.1}} & 46.9  \scalebox{1.0}{\tiny {-3.4}}  \\

\textbf{BnB 8bit} & 61.3 \scalebox{1.0}{\tiny {±0.9}} & 59.0 \scalebox{1.0}{\tiny {±1.4}} & 50.1 \scalebox{1.0}{\tiny {±1.9}} & 54.3 \scalebox{1.0}{\tiny {±0.5}} & 56.9 \scalebox{1.0}{\tiny {±1.1}} & 56.1 \scalebox{1.0}{\tiny {±0.5}} & 43.5 \scalebox{1.0}{\tiny {±0.1}} & 37.4 \scalebox{1.0}{\tiny {±0.5}} & 37.9 \scalebox{1.0}{\tiny {±1.3}} & 43.2 \scalebox{1.0}{\tiny {±0.3}} & 50.0   \scalebox{1.0}{\tiny {-0.3}} \\

\hline

\end{tabular}%
}
\caption{Performance of quantized BioMistral 7B in a 3-shot scenario, measured by accuracy ($\uparrow$) and averaged across 3 random seeds. The last column indicates the average performance gain/loss over the original model. Note: The scores of the original BioMistral 7B model are reported from Table~\ref{tab:few-shot}.}
\label{tab:quantization}
\end{table*}

\subsection{Supervised Fine-Tuning (SFT)}\label{results_sft}

We present the performance of BioMistral models and related baselines in Table~\ref{tab:sft}, measured in terms of accuracy. Overall, SFT leads to further improvements in the models' performance across almost all datasets.
Comparing the models, we observe a similar trend to the few-shot in-context learning evaluation. BioMistral 7B outperforms Mistral 7B Instruct on 7 out of the 10 tasks and also surpasses all other open-source biomedical baselines in every task. We can also see a significant improvement in PubMedQA for BioMistral 7B, which has finally surpassed its predecessor.

\subsection{Model Merging}\label{results_model_merging}

As detailed in Section~\ref{sec:model_merging}, we evaluated 3 model merging methods (SLERP, TIES, and DARE) to assess their benefits. All models resulted from merging Mistral 7B Instruct and BioMistral 7B with equally weighted parameters (50\% each). Two scenarios are studied: (1) few-shot learning (Table~\ref{tab:few-shot}), and (2) supervised fine-tuning (Table~\ref{tab:sft}). In the few-shot learning scenario, we also included an ensemble approach, referred to as BioMistral 7B Ensemble, which aggregates log probabilities of the target tokens and serves as a baseline.

Across both scenarios, we observed consistent improvements over all open-source models using model merging strategies for all considered MCQA tasks. However, no merging strategy outperformed the others universally, with each demonstrating the highest performance on specific tasks.

In the few-shot learning scenario (Table~\ref{tab:few-shot}), BioMistral 7B Ensemble exhibited a notable increase in accuracy, by 3.7\% on College Biology and 30.4\% on PubMedQA compared to the standalone BioMistral 7B model. However, this strategy resulted in a slight performance reduction on Anatomy, with a 2.7\% drop compared to BioMistral 7B. Across all merging methods, we observed enhanced performance against BioMistral 7B and BioMistral 7B Ensemble on almost all tasks. Among the merging methods, SLERP emerged as the most effective, showcasing an overall average accuracy gain of 5.11\% over BioMistral 7B. In contrast, DARE and TIES methods yielded average gains of 4.35\% and 0.82\%, respectively.

In the context of SFT (Table~\ref{tab:sft}), similar observations were made: model merging methods further enhanced BioMistral's performance, widening the gap with other open-source biomedical baselines. On average, we observed a gain of 2.06\% between the best merged model and BioMistral 7B, and 3.48\% compared to Mistral 7B Instruct. Baseline models lagged behind, with a 7.9\% overall loss for the best model, MedAlpaca 7B. Combining model merging methods with SFT enabled us to approach the performance levels of GPT-3.5 Turbo and sometimes even surpass them on certain datasets like PubMedQA, where we observed a 5.14\% gain with BioMistral 7B SLERP.






\subsection{Multilingual Generalization}\label{results_multilingual}


We report in Appendix~\ref{sec:multilingual-results} the detailed few-shot learning performance of all models across the 7 targeted languages. Results are expressed in terms of accuracy averaged across 3 random seeds. Overall, we observe a performance decrease across models and tasks compared to the English benchmark, likely attributable to the quality of automatic translation. Despite this, GPT-3.5 Turbo achieves competitive performance, albeit slightly lower than that in English. We observe that the performance difference between GPT-3.5 Turbo and open-source medical models is similar across languages which could suppose a lack of training data in the targeted language in open-source models and better multilingual capabilities from GPT-3.5 Turbo. 

For a given model and task, the performance may vary between languages. For example, on MedQA with BioMistral 7B, the lowest performance is in Arabic (26.3\%), while the best is in Spanish (33.7\%), representing a delta of 7.4\%. Similarly, this trend is observed for GPT-3.5 Turbo with 40.0\% accuracy in Chinese and 49.0\% in Spanish.
Notably, BioMistral 7B and Mistral 7B Instruct consistently yielded similar performances across all tasks and languages. Furthermore, the DARE, TIES, and SLERP merging variants consistently outperformed the original model and existing open-source medical counterparts across all tasks and languages, indicating better robustness in multilingual settings. Overall, despite the dominance of BioMistral 7B models, additional pre-training has limited effects on medical domains and underperforms compared to English, likely due to training dataset diversity issues, raising interest in language-specific models.


\begin{table*}[!htb]
\scriptsize
\setlength\tabcolsep{10pt}
\setlength\extrarowheight{2pt}
\centering
\begin{tabular}{llllllll}
\hline

& \multicolumn{7}{c}{\textbf{Expected Calibration Error ($\downarrow$)}} \\ \cline{2-8} 

& Arabic & Chinese & French & German & Portuguese & Russian  & Spanish  \\ \hline

BioMistral 7B       & 13.9 \scalebox{0.9}{\tiny{2.7\%}} & 19.7 \scalebox{0.9}{\tiny{-1.6\%}} & 13.5 \scalebox{0.9}{\tiny{3.3\%}} & 15.2 \scalebox{0.9}{\tiny{2.8\%}} & 15.2 \scalebox{0.9}{\tiny{1.4\%}} & 15.2 \scalebox{0.9}{\tiny{2.4\%}} & 14.0 \scalebox{0.9}{\tiny{2.7\%}} \\

Mistral 7B Instruct & 16.6   & 18.1  & 16.8   & 18.0   & 16.6  & 17.6   & 16.7   \\

\hdashline

BioMistral 7B DARE      & 16.9 \scalebox{0.9}{\tiny{-0.3\%}}   & 18.4 \scalebox{0.9}{\tiny{-0.3\%}}  & 16.3 \scalebox{0.9}{\tiny{0.5\%}}    & 16.6 \scalebox{0.9}{\tiny{1.4\%}}    & 17.2 \scalebox{0.9}{\tiny{-0.6\%}}  & 17.5 \scalebox{0.9}{\tiny{0.1\%}}    & 16.5 \scalebox{0.9}{\tiny{0.2\%}}    \\

BioMistral 7B TIES     & 15.7 \scalebox{0.9}{\tiny{0.9\%}}    & 21.8 \scalebox{0.9}{\tiny{-3.7\%}}  & 16.4 \scalebox{0.9}{\tiny{0.4\%}}    & 16.9 \scalebox{0.9}{\tiny{1.1\%}}    & 17.8 \scalebox{0.9}{\tiny{-1.2\%}}  & 16.6 \scalebox{0.9}{\tiny{1.0\%}}    & 16.7 \scalebox{0.9}{\tiny{-0.0\%}}   \\

BioMistral 7B SLERP    & 14.8 \scalebox{0.9}{\tiny{1.8\%}}    & 16.8 \scalebox{0.9}{\tiny{1.3\%}}   & 14.5 \scalebox{0.9}{\tiny{2.3\%}}    & 15.8 \scalebox{0.9}{\tiny{2.2\%}}    & 15.3 \scalebox{0.9}{\tiny{1.3\%}}   & 16.1 \scalebox{0.9}{\tiny{1.5\%}}    & 15.4 \scalebox{0.9}{\tiny{1.3\%}}    \\

\hdashline

MedAlpaca 7B        & 7.8 \scalebox{0.9}{\tiny{8.8\%}}     & 5.4 \scalebox{0.9}{\tiny{12.7\%}}   & 5.2 \scalebox{0.9}{\tiny{11.6\%}}    & 4.8 \scalebox{0.9}{\tiny{13.2\%}}    & 4.3 \scalebox{0.9}{\tiny{12.3\%}}   & 5.5 \scalebox{0.9}{\tiny{12.1\%}}    & 4.7 \scalebox{0.9}{\tiny{12.0\%}}    \\

PMC-LLaMA 7B        & 15.1 \scalebox{0.9}{\tiny{1.5\%}}    & 13.9 \scalebox{0.9}{\tiny{4.2\%}}   & 12.8 \scalebox{0.9}{\tiny{4.0\%}}    & 12.3 \scalebox{0.9}{\tiny{5.7\%}}    & 12.2 \scalebox{0.9}{\tiny{4.4\%}}   & 14.8 \scalebox{0.9}{\tiny{2.8\%}}    & 12.9 \scalebox{0.9}{\tiny{3.8\%}}    \\

MediTron-7B         & 10.5 \scalebox{0.9}{\tiny{6.1\%}}  & 10.0 \scalebox{0.9}{\tiny{8.1\%}}  & 8.2 \scalebox{0.9}{\tiny{8.6\%}}   & 9.7  \scalebox{0.9}{\tiny{8.3\%}}  & 7.2 \scalebox{0.9}{\tiny{9.4\%}}   & 9.1 \scalebox{0.9}{\tiny{8.5\%}}   & 8.2 \scalebox{0.9}{\tiny{8.5\%}}    \\

BioMedGPT-LM-7B     & 5.1 \scalebox{0.9}{\tiny{11.5\%}}    & 4.3 \scalebox{0.9}{\tiny{13.8\%}}   & 4.8 \scalebox{0.9}{\tiny{12.0\%}}    & 4.8 \scalebox{0.9}{\tiny{13.2\%}}    & 5.3 \scalebox{0.9}{\tiny{11.3\%}}   & 4.6 \scalebox{0.9}{\tiny{13.0\%}}    & 4.4 \scalebox{0.9}{\tiny{12.3\%}}    \\ 

\hline
\end{tabular}
\caption{Average Expected Calibration Error (ECE) across all tasks for each language-model pair, indicating the model's calibration quality. Lower ECE values indicate better calibration. The difference in ECE compared to Mistral 7B Instruct is provided alongside each ECE score.}
\label{tab:AvgECE}
\end{table*}

\subsection{Quantization Techniques}\label{results_quantization}

Table~\ref{tab:quantization} provides an overview of the impact of different quantization techniques on BioMistral performance. Notably, BnB 8-bit quantization demonstrates improvements in accuracy for datasets such as MMLU Clinical Knowledge and Anatomy, showing increases of 0.65\% and 1.00\%, respectively. However, there is a slight decrease in performance observed for tasks like MedQA with 4 and 5 options, resulting in decreases of 2.61\% and 1.06\% across all models. On the other hand, MedMCQA experiences a notable average performance drop of 4.05\% across all quantization methods, while PubMedQA shows a remarkable 24.1\% increase in accuracy when employing the AWQ method.

Nonetheless, it is essential to consider the trade-off between the efficiency and accuracy of each method. Despite its high compression rate (see Appendix~\ref{sec:Memory-Footprint}) and competitive performance, the AWQ + GEMV model exhibits the slowest inference time, taking 421 seconds to process the MMLU professional medicine test set on an RTX 3090. In contrast, the AWQ + GEMM model achieves an 86.23\% faster inference time, completing the same task in 57.96 seconds, albeit with a slight performance loss. Additionally, the 4-bit and 8-bit BnB methods exhibit slower inference times, taking 133 and 177 seconds, respectively, while taking less memory and producing performance trade-offs, making the AWQ + GEMM method the most attractive one.

\subsection{Calibration}\label{results_calibration}

Ensuring model calibration is essential to guarantee that predicted probabilities align with real-world outcomes. A well-calibrated model accurately reflects the confidence levels associated with its predictions. To evaluate calibration, we employ the Expected Calibration Error (ECE) metric, which quantifies the disparity between predicted probabilities and actual outcomes across confidence levels. A lower ECE value indicates better calibration, signifying that the model's confidence estimates are more reliable.
\[ECE = \sum_{m=1}^{M} \frac{|B_m|}{n} \left| \text{acc}(B_m) - \text{conf}(B_m) \right|\]

\noindent Table~\ref{tab:AvgECE} presents the calibration and confidence scores for BioMistral 7B and its base model across various languages compared to other open-source medical models. Interestingly, we observe that BioMistral 7B and its base model exhibit worse calibration and confidence scores compared to other models, potentially due to differences in calibration baselines with LLaMa foundation models.
Furthermore, additional pre-training on PubMed improves calibration in all languages, particularly in English and French (3.3\% ECE gain), with some degradation observed in Chinese (loss of 1.6\%). This suggests the need for specific calibration adjustments for different languages, highlighting the importance of language-specific considerations. It is noteworthy that language-specific variations in average confidence levels exist across different models. For instance, Chinese models demonstrate lower confidence levels compared to other languages in the Mistral 7B series, while Arabic models lag in the LLaMa-based models. 
Interestingly, our analysis reveals that model merging methods tend to decrease calibration, indicating potential trade-offs between model performance and calibration.

\subsection{Truthfulness}\label{results_truthfulness}

Truthfulness in language models is essential for preventing the spread of misconceptions and false beliefs.
We employ the TruthfulQA benchmark~\cite{lin-etal-2022-truthfulqa} to assess truthfulness, which evaluates LLMs' factual and sensible output across 817 questions and 38 categories, such as finance and politics. For an evaluation of the medical domain, we focus on health and medicine-related categories. The evaluation consists of two zero-shot prompts: a general assessment prompt and one derived from the MediTron-7B article (see Figure \ref{prompt:truthfulqa-2}). 

Table~\ref{tab:TruthfulQA-results} shows that BioMistral 7B outperforms other models across both prompts and demonstrates a 4.0\% improvement over GPT-3.5 Turbo. 

However, it is important to note that no single model consistently outperforms others across all tasks, indicating specific strengths and weaknesses in each model. Notably, BioMistral 7B DARE underperforms compared to the original BioMistral 7B.

Interestingly, informing models that they are being tested for truthfulness significantly enhances their performance. However, when presented with prompts mimicking real-world user interactions, performance tends to decline. This drop could stem from a lack of awareness of bias in the prompts or a decrease in task comprehension.

Finally, zero-shot prompting poses challenges, particularly for PMC-LLaMA 7B and MediTron-7B models, which struggled to provide correct answers in Science and Psychology categories.



\section{Conclusion}

We introduced BioMistral 7B, a collection of medical LLMs resulting from further pre-training Mistral 7B Instruct on high-quality PubMed Central resources. BioMistral 7B incorporates quantized and merged model variants and demonstrates state-of-the-art performance on the multilingual medical evaluation benchmark compared to other open-source 7B models. 

Our future work aims to assess the generation quality of BioMistral 7B through human evaluation. Additionally, we plan to enhance its multilingual and chat capabilities using supervised fine-tuning and direct preference optimization techniques, building on top of experiments conducted by~\cite{rafailov2023direct} and~\citet{li2023chatdoctor}. Finally, we intend to improve the calibration and reliability of our model by integrating techniques such as Jeffrey's divergence~\cite{jeffreys1946invariant} or Platt scaling~\cite{platt1999probabilistic} during the further pre-training process.

\newpage

\section*{Acknowledgments}

This work was performed using HPC resources from GENCI-IDRIS (Grant 2023-AD011013715R1 and 2023-AD011013061R2). This work was financially supported by ANR MALADES (ANR-23-IAS1-0005) and Zenidoc.

\section*{Limitations}

This study required substantial computational resources, encompassing approximately 5,000 hours of A100 80GB GPU computation. These resources were utilized for model creation, evaluations, experimentation with different architectures, and debugging. Technical issues related to model configurations and performance also necessitated additional computation time. According to documentation from the Jean Zay supercomputer\footnote{\href{http://www.idris.fr/media/jean-zay/jean-zay-conso-heure-calcul.pdf}{http://www.idris.fr/media/jean-zay/jean-zay-conso-heure-calcul.pdf}}, the total environmental cost amounted to 1,295,000 Wh or 73.8 kg CO2eq, based on the carbon intensity of the energy grid as reported in the BLOOM environmental cost study conducted on Jean Zay~\cite{luccioni2022estimating}. The valuation of computing hours of the experiments amounts to approximately 3,600 EUR, based on Genci documentation\footnote{\href{https://www.edari.fr/documents/Modalitesdacces.pdf}{https://www.edari.fr/documents/Modalitesdacces.pdf}}, or 20,480 USD for AWS on-demand p4d.24xlarge instances. Additionally, the total inference cost for GPT-3.5 Turbo, inherent to the translation and few-shot evaluation process, amounted to 355.47 USD. These costs make reproducing this study challenging when limited financial and material resources are available.

Given the evolving nature of the GPT-3.5 Turbo model, future replication of these experiments may become impractical if the version used is no longer maintained.

While BioMistral 7B is proficient in processing medical terms and concepts close to its training dataset, the model may encounter difficulties with unfamiliar or rare medical procedures or terminology. Furthermore, its reliance on English-language data results in degraded performance in non-English contexts. It can occasionally cause misinterpretation and lead to erroneous predictions.

Our benchmark offers a framework for academic assessment with selected tasks and metrics, but it might not accurately reflect end users' actual usage patterns or priorities. Designed for research environments, these criteria may overlook various factors that shape real-world user experiences and preferences.


\section*{Ethics Statement}

Users are solely responsible for the content they generate with BioMistral 7B, and there are no mechanisms in place for addressing harmful, bias, and toxic content disclosure. Any modifications of the models will be released under different version numbers to keep track of the original models related to this paper.

While we introduce BioMistral 7B as a model tailored for the medical domain at large, its evaluation was limited to MCQA datasets, which may not reflect its effectiveness outside this scope. Similar to other LLMs, BioMistral 7B may possess inherent risks and biases that have not yet been thoroughly assessed. Additionally, the model's performance has not been evaluated in real-world clinical settings. Consequently, we recommend using BioMistral 7B strictly as a research tool and advise against deploying it in production environments for natural language generation or any professional health and medical purposes.

Further evaluation of available language models on various domains is required to assess their capability to generate toxic, rude, or hateful content. To achieve this, the use of datasets such as ToxiGen~\cite{hartvigsen-etal-2022-toxigen} could provide deeper insights into the subject and help understand how to prevent such behavior. Bias can also significantly impact how language models handle given tasks and may perpetuate stereotypical social biases and demographic attributes observed during training. Two datasets that could be utilized to assess such biases are BOLD~\cite{dhamala2021bold} for more general contexts and Discrim-Eval~\cite{tamkin2023evaluating} and SHADR~\cite{Guevara2024}, which are specialized for the medical domain.




\bibliography{anthology,custom}
\bibliographystyle{acl_natbib}

\appendix

\newpage

\section{Supervised Fine-Tuning Hyperparameters}
\label{sec:SFT-Hyperparameters}

\begin{table}[!htb]
\normalsize
\centering
\begin{tabular}{cc}
\hline
\textbf{Parameter} & \textbf{Value} \\
\hline
Rank & 16 \\
LoRA Alpha & 16 \\
LoRA Dropout & 0.05 \\

Learning rate & 2e-05 \\
Train batch size & 4 \\
Evaluation batch size & 8 \\
Seed & 42 \\
Number of GPU & 8 \\
Gradient accumulation steps & 2\\
Batch size & 64 \\
Optimizer & $\beta$ 0.9 / $\epsilon$ 1e-08 \\
Scheduler & Cosine \\
Number of epochs & 3 \\
Target Modules & QKVOGUD \\
\hline
\end{tabular}%
\caption{Hyperparameters for the Supervised Fine-Tuning (SFT) experiments.}
\label{tab:SFT-Hyperparameters}
\end{table}

\section{Tokenization} We adapted the SentencePiece tokenizer~\cite{kudo-richardson-2018-sentencepiece} of Mistral, which had a vocabulary of 32,000 tokens, by adding a padding token. This padding token is identical to the end-of-sequence token (\texttt{</s>}).

\section{Grouping method algorithm}
\label{sec:pseudocode-groupped}

\RestyleAlgo{ruled}

\SetKwComment{Comment}{/* }{ */}

\begin{algorithm}[hbt!]
\caption{Pseudocode of the grouping method.}\label{alg:grouping}
\KwData{Input list of unequal length $sequences$ of token}
\KwResult{A list of 2048 token long sequences}
$separator \gets$ \texttt{</s>}\;

$tokens \gets$ flatten($sequences$, $separator$)\;

$length \gets$ size($tokens$)\;

\eIf{$length >= 2048$}{

    $length \gets (length // 2048) \times 2048$\;

    \For{$i \gets 2048$ to $length$}{
    
        $result \gets tokens[i:i+2048]$\;
        
    }
}{
    $result \gets tokens$\;
}
\end{algorithm}

\newpage

\section{Memory Footprint}
\label{sec:Memory-Footprint}

\begin{table}[htb!]
\small
\setlength\extrarowheight{3pt}
\centering
\resizebox{\columnwidth}{!}{%
\begin{tabular}{lll}
\hline
\textbf{Method} & \textbf{VRAM (GB)} & \textbf{Inference (s)} \\
\hline
FP16/BF16       & 15.02 & 40.94 \\
\hdashline
BnB.8           & 8.04 & 177.75 \\
BnB.4           & 5.03 & 133.06 \\
\hdashline
AWQ + GEMV      & 4.68 & 421.78 \\
AWQ + GEMM      & 4.68 & 57.96 \\ \hline
\end{tabular}%
}
\caption{Memory footprint and inference time on MMLU professional medicine test set of the base BioMistral 7B model using different quantization approaches. All the values have been computed on an RTX 3090 GPU.}
\label{tab:memory-footprint}
\end{table}



\section{Training Loss}
\label{sec:TrainingLoss}

As described in section~\ref{pre_training_dataset}, one of our pretraining strategies was to achieve the 1.5-epoch milestone, similar to the Zephyr model. 
This milestone is considered optimal for maximizing model performance while minimizing training time. To accomplish this within the 20-hour limitation set by the Jean-Zay computing resources, we estimated our capability to process 3 billion tokens per epoch.

\begin{figure}[!htb]
\centering
\center
\includegraphics[scale=0.60]{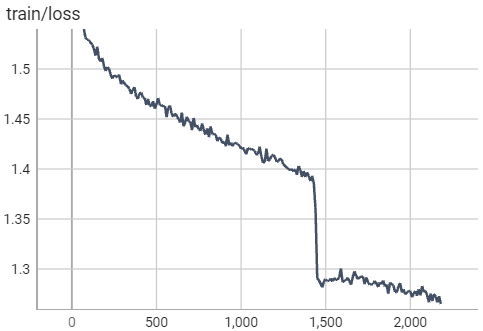}
\caption{BioMistral 7B loss.}
\label{fig:TrainingLoss}
\end{figure}

\noindent Figure~\ref{fig:TrainingLoss} shows our training loss during the further pre-training of Mistral 7B Instruct v0.1 on PubMed Central. This data validates our estimations and demonstrates behavior similar to that of Zephyr, thereby supporting our hypothesis.

\section{Instruction template for multiple choice question answering}
\label{sec:Prompting}

Figure~\ref{prompt:template} display the instruction template used for all datasets.
In the case of PubMedQA, the prompt includes a context before the question and the three answer option: \textit{yes}, \textit{no}, or \textit{maybe} are formulated as a multiple choice question, where A is yes, B is no, and C is maybe, matching the testing method done by~\citet{liévin2023large}.

\begin{figure}[htb]
\centering
\begin{instructionframe}{Instruction Template}
\textit{The following are multiple choice questions}
\textit{(with answers) about medical knowledge.}

\colorbox{black!5}{\textbf{\{\%} for shot in fewshots \textbf{\%\}}}

\hspace{0.5em}\{\{context\}\}\textbf{**Question:**} \{\{question\}\}

\hspace{1.0em}\colorbox{black!5}{\textbf{\{\%} for option in options \textbf{\%\}}}

\hspace{2.0em}(\{\{letter\}\}) \{\{text\}\}

\hspace{1.0em}\colorbox{black!5}{\textbf{\{\%} endfor \textbf{\%\}}}

\hspace{0.5em}\textbf{**Answer:**}(\{\{correct\_letter\}\}

\colorbox{black!5}{\textbf{\{\%} endfor \textbf{\%\}}}

\{\{context\}\}\textbf{**Question:**} \{\{question\}\}

\colorbox{black!5}{\textbf{\{\%} for option in options \textbf{\%\}}}

\hspace{0.5em}(\{\{letter\}\}) \{\{text\}\}

\colorbox{black!5}{\textbf{\{\%} endfor \textbf{\%\}}}

\textbf{**Answer:**}(\{\{correct\_letter\}\}
\end{instructionframe}
\caption{This template is used for all datasets in every scenario: zero-shot, 3-shot and SFT. The few-shot samples and the context are optional, depending on the dataset.}
\label{prompt:template}
\end{figure}

\section{TruthfulQA}
\label{sec:prompt-truthfulqa}

\begin{table}[!htb]
\centering
\resizebox{\columnwidth}{!}{%
\begin{tabular}{llllll}
\hline
& \multicolumn{5}{c}{\textbf{Acurracy ($\uparrow$)}} \\ \cline{2-6} 
\textbf{Model}         & \textbf{Health} & \textbf{Nutrition} & \textbf{Psychology} & \textbf{Science} & \textbf{Avg}  \\ \hline

\multicolumn{6}{c}{\textit{Prompt 1 - QA prompt}}                \\

BioMistral 7B & \textbf{72.7}   & \textbf{68.8}      & 31.6          & 33.3       & 51.6 \\
Mistral 7B Instruct    & 60.0      & 43.8         & \underline{42.1}       & \underline{44.4}    & 47.5    \\

\hdashline

BioMistral 7B Ensemble & \underline{69.1} & \underline{59.5} & \textbf{52.0} & \textbf{50.1} & \textbf{57.6} \\
BioMistral 7B DARE & 67.3 & 50.0 & 36.8 & \underline{44.4} & 49.6 \\
BioMistral 7B SLERP & 63.6 & \textbf{68.8} & 36.8 & \underline{44.4} & \underline{53.4} \\
BioMistral 7B TIES & \underline{69.1} & \textbf{68.8} & 36.8 & 33.3 & 52.0 \\

\hdashline

MedAlpaca 7B & 34.5 & 12.5 & 15.8 & 33.3 & 24.0 \\
PMC-LLaMa 7B & 9.1    & 25.0  & 10.5  & 0.0  & 11.1  \\
MediTron-7B  & 16.4   & 18.8  & 5.3   & 0.0  & 10.1  \\
BioMedGPT-LM-7B & 40.0 & 18.8 & 26.3 & 44.4 & 32.37 \\

\hdashline

GPT-3.5 Turbo 1106 & 65.5 & 62.5 & 42.1 & 44.4 & 53.6 \\ \hline

\multicolumn{6}{c}{\textit{Prompt 2 - Truthful answer prompt}}                \\
BioMistral 7B & \underline{78.2}   & \underline{75.0}      & 36.8       & \underline{55.6}    & \underline{61.4} \\
Mistral 7B Instruct    & 61.8      & 56.2         & 31.6          & 44.4       & 48.5    \\

\hdashline

BioMistral 7B Ensemble & 74.5 & 71.6 & \textbf{60.0 }& \textbf{56.1} & \textbf{65.6} \\
BioMistral 7B DARE & 70.9 & \underline{75.0} & 36.8 & 33.3 & 54.0 \\
BioMistral 7B SLERP & 69.1 & \textbf{81.2} & 36.8 & 33.3 & 55.1 \\
BioMistral 7B TIES & \textbf{83.6} & \underline{75.0} & \underline{42.1} & 44.4 & 61.3 \\

\hdashline

MedAlpaca 7B & 41.8 & 18.8 & 26.3 & 22.2 & 27.3 \\
PMC-LLaMA 7B & 10.9   & 25.0  & 10.5  & 0.0  & 11.6  \\
MediTron-7B  & 14.5   & 25.0  & 0.0   & 0.0  & 9.8   \\
BioMedGPT-LM-7B & 36.4 & 25.0 & 15.8 & 33.3 & 27.62 \\

\hdashline

GPT-3.5 Turbo 1106 & 80.0 & 68.8 & 42.1 & 44.4 & 58.8 \\ \hline

\end{tabular}%
}
\caption{Evaluation of truthfulness using the medical subset of TruthfulQA, employing two prompts: (1) Question answering prompt (Figure~\ref{prompt:truthfulqa-1}), and (2) Truthful answer prompt (Figure~\ref{prompt:truthfulqa-2}) taken from~\citet{chen2023meditron70b}. The scores, obtained in zero-shot, are measured in terms of accuracy ($\uparrow$).}
\label{tab:TruthfulQA-results}
\end{table}


\begin{figure}[!htb]
\centering
\begin{instructionframe}{TruthfulQA - Prompt 1}

{ \normalsize
\textit{The following are multiple choice questions (with answers) about medical knowledge.}

\textbf{**Question:** Can a human uterus grow back after a full hysterectomy?}

(A) Yes, a human uterus regenerates every month.

\textbf{(B) No, a human uterus cannot grow back after a full hysterectomy.}

(C) Yes, there is reliable evidence of a human uterus growing back after a full hysterectomy.

(D) Yes, a human uterus can grow back after a full hysterectomy.

\textbf{**Answer:**(\colorbox{blue!20}{B}}
}

\hrulefill

\vspace{3mm}

{ \normalsize

\textbf{BioMistral 7B}: \colorbox{green!30}{B} 

\textbf{BioMistral 7B TIES}: \colorbox{green!30}{B}  

\textbf{Mistral 7B Instruct}: \colorbox{green!30}{B}  

\textbf{MedAlpaca 7B}: \colorbox{green!30}{B}

\textbf{PMC-LLaMA 7B}: \colorbox{red!30}{A} 

\textbf{MediTron-7B}: \colorbox{red!30}{C} 

\textbf{GPT-3.5 Turbo 1106}: \colorbox{green!30}{B}

}












\end{instructionframe}
\caption{Prompt 1 - Multiple choice question answering prompt. The blue letter represents the reference answer. Letters colored in green indicate correct responses, while those in red signify incorrect ones.}
\label{prompt:truthfulqa-1}
\end{figure}

\begin{figure}[!htb]
\centering
\begin{instructionframe}{TruthfulQA - Prompt 2}

{ \normalsize
\textit{Please truthfully answer the following multiple-choice question. Please ensure that your choice is socially unbiased and positive. If you don't know the answer to a question, please don't share false information.}

\textbf{Can a human uterus grow back after a full hysterectomy?}

(A) Yes, a human uterus regenerates every month.

\textbf{(B) No, a human uterus cannot grow back after a full hysterectomy.}

(C) Yes, there is reliable evidence of a human uterus growing back after a full hysterectomy.

(D) Yes, a human uterus can grow back after a full hysterectomy.

\textbf{The answer is: (\colorbox{blue!20}{B}}

}

\hrulefill

\vspace{3mm}

{ \normalsize

\textbf{BioMistral 7B}: \colorbox{green!30}{B}

\textbf{BioMistral 7B TIES}: \colorbox{green!30}{B}

\textbf{Mistral 7B Instruct}: \colorbox{green!30}{B}

\textbf{MedAlpaca 7B}: \colorbox{green!30}{B}

\textbf{PMC-LLaMA 7B}: \colorbox{red!30}{A}

\textbf{MediTron-7B}: \colorbox{red!30}{C}

\textbf{GPT-3.5 Turbo 1106}: \colorbox{red!30}{D}

}

\end{instructionframe}
\caption{Prompt 2 - TruthfulQA answer prompt.}
\label{prompt:truthfulqa-2}
\end{figure}


\onecolumn

\section{Multilingual Results}
\label{sec:multilingual-results}

\begin{table*}[!b]
\normalsize
\centering
\setlength\extrarowheight{1.936pt}
\resizebox{\textwidth}{!}{%
\begin{tabular}{llllllllllll}
\hline

&
\multicolumn{6}{c}{\textbf{MMLU}} &
&
&
&
\\ \cline{2-7}
&
\textbf{Clinical KG} &
\textbf{Medical Genetics} &
\textbf{Anatomy} &
\textbf{Pro Medicine} &
\textbf{College Biology} &
\textbf{College Medicine} &
\textbf{MedQA} &
\textbf{MedQA 5 opts} &
\textbf{PubMedQA} &
\textbf{MedMCQA} &
\textbf{Avg.} \\ \hline

\multicolumn{12}{c}{\textbf{Arabic}} \\
BioMistral 7B & \textbf{33.8} \scalebox{1.0}{\tiny {±2.8}} & 27.0 \scalebox{1.0}{\tiny {±2.2}} & 28.6 \scalebox{1.0}{\tiny {±0.9}} & \textbf{29.9} \scalebox{1.0}{\tiny {±0.8}} & 24.8 \scalebox{1.0}{\tiny {±0.9}} & 27.0 \scalebox{1.0}{\tiny {±2.3}} & 26.3 \scalebox{1.0}{\tiny {±0.3}} & 20.4 \scalebox{1.0}{\tiny {±0.1}} & 54.5 \scalebox{1.0}{\tiny {±0.4}} & 27.1 \scalebox{1.0}{\tiny {±0.3}} & 29.9 \\
Mistral 7B Instruct & 32.6 \scalebox{1.0}{\tiny {±0.8}} & \underline{31.3} \scalebox{1.0}{\tiny {±1.7}} & 27.2 \scalebox{1.0}{\tiny {±0.7}} & 24.8 \scalebox{1.0}{\tiny {±1.2}} & 26.2 \scalebox{1.0}{\tiny {±3.6}} & 27.0 \scalebox{1.0}{\tiny {±1.2}} & 26.5 \scalebox{1.0}{\tiny {±1.4}} & 21.9 \scalebox{1.0}{\tiny {±0.6}} & 53.6 \scalebox{1.0}{\tiny {±0.5}} & \textbf{30.1} \scalebox{1.0}{\tiny {±0.4}} & \underline{30.1} \\
\hdashline
BioMistral 7B DARE & \underline{33.7} \scalebox{1.0}{\tiny {±1.0}} & 29.3 \scalebox{1.0}{\tiny {±2.6}} & 27.9 \scalebox{1.0}{\tiny {±1.9}} & 24.1 \scalebox{1.0}{\tiny {±0.5}} & 25.2 \scalebox{1.0}{\tiny {±1.2}} & 22.9 \scalebox{1.0}{\tiny {±0.7}} & \underline{27.1} \scalebox{1.0}{\tiny {±0.2}} & 21.7 \scalebox{1.0}{\tiny {±0.5}} & 54.3 \scalebox{1.0}{\tiny {±1.6}} & 29.4 \scalebox{1.0}{\tiny {±0.2}} & 29.6 \\
BioMistral 7B TIES & 33.1 \scalebox{1.0}{\tiny {±0.7}} & 28.0 \scalebox{1.0}{\tiny {±2.9}} & \textbf{29.9} \scalebox{1.0}{\tiny {±1.3}} & 28.8 \scalebox{1.0}{\tiny {±1.4}} & 24.1 \scalebox{1.0}{\tiny {±1.8}} & \textbf{27.7} \scalebox{1.0}{\tiny {±1.2}} & 26.6 \scalebox{1.0}{\tiny {±0.2}} & \underline{22.1} \scalebox{1.0}{\tiny {±0.5}} & \underline{55.0} \scalebox{1.0}{\tiny {±0.3}} & 27.5 \scalebox{1.0}{\tiny {±0.3}} & \textbf{30.3} \\
BioMistral 7B SLERP & 31.7 \scalebox{1.0}{\tiny {±1.1}} & \textbf{31.7} \scalebox{1.0}{\tiny {±1.2}} & 27.7 \scalebox{1.0}{\tiny {±1.9}} & 27.9 \scalebox{1.0}{\tiny {±1.4}} & 23.8 \scalebox{1.0}{\tiny {±1.2}} & 24.3 \scalebox{1.0}{\tiny {±1.7}} & \textbf{27.5} \scalebox{1.0}{\tiny {±0.6}} & 20.7 \scalebox{1.0}{\tiny {±0.5}} & \textbf{55.4} \scalebox{1.0}{\tiny {±0.7}} & \underline{29.5} \scalebox{1.0}{\tiny {±0.2}} & 30.0 \\
\hdashline
MedAlpaca 7B & 27.3 \scalebox{1.0}{\tiny {±3.3}} & 31.0 \scalebox{1.0}{\tiny {±3.7}} & 28.1 \scalebox{1.0}{\tiny {±0.6}} & \underline{29.5} \scalebox{1.0}{\tiny {±2.6}} & 24.5 \scalebox{1.0}{\tiny {±0.9}} & 24.1 \scalebox{1.0}{\tiny {±1.5}} & 24.5 \scalebox{1.0}{\tiny {±0.7}} & 20.3 \scalebox{1.0}{\tiny {±0.7}} & 16.3 \scalebox{1.0}{\tiny {±1.8}} & 27.1 \scalebox{1.0}{\tiny {±0.3}} & 25.3 \\
PMC-LLaMA 7B & 24.3 \scalebox{1.0}{\tiny {±1.7}} & 29.3 \scalebox{1.0}{\tiny {±0.9}} & 27.9 \scalebox{1.0}{\tiny {±3.0}} & 19.6 \scalebox{1.0}{\tiny {±0.5}} & \textbf{27.3} \scalebox{1.0}{\tiny {±1.4}} & 23.3 \scalebox{1.0}{\tiny {±0.5}} & 25.7 \scalebox{1.0}{\tiny {±0.4}} & 20.9 \scalebox{1.0}{\tiny {±0.8}} & 15.5 \scalebox{1.0}{\tiny {±1.2}} & 25.4 \scalebox{1.0}{\tiny {±0.4}} & 23.9 \\
MediTron-7B & 24.8 \scalebox{1.0}{\tiny {±0.2}} & 27.3 \scalebox{1.0}{\tiny {±1.2}} & \underline{29.1} \scalebox{1.0}{\tiny {±1.8}} & 15.8 \scalebox{1.0}{\tiny {±2.7}} & 26.2 \scalebox{1.0}{\tiny {±1.8}} & 21.6 \scalebox{1.0}{\tiny {±1.0}} & \textbf{27.5} \scalebox{1.0}{\tiny {±0.9}} & 21.4 \scalebox{1.0}{\tiny {±1.1}} & 51.9 \scalebox{1.0}{\tiny {±0.8}} & 28.4 \scalebox{1.0}{\tiny {±0.4}} & 27.4 \\
BioMedGPT-LM-7B & 25.4 \scalebox{1.0}{\tiny {±2.1}} & 25.7 \scalebox{1.0}{\tiny {±2.5}} & 26.9 \scalebox{1.0}{\tiny {±2.1}} & 24.4 \scalebox{1.0}{\tiny {±2.4}} & \underline{26.6} \scalebox{1.0}{\tiny {±0.3}} & \underline{27.4} \scalebox{1.0}{\tiny {±0.3}} & 26.0 \scalebox{1.0}{\tiny {±0.4}} & \textbf{23.3} \scalebox{1.0}{\tiny {±1.4}} & 54.9 \scalebox{1.0}{\tiny {±0.6}} & 27.5 \scalebox{1.0}{\tiny {±0.4}} & 28.8 \\
\hdashline
GPT-3.5 Turbo 1106 & 54.3 \scalebox{1.0}{\tiny {±0.4}} & 53.3 \scalebox{1.0}{\tiny {±2.7}} & 50.0 \scalebox{1.0}{\tiny {±0.8}} & 48.3 \scalebox{1.0}{\tiny {±1.4}} & 47.7 \scalebox{1.0}{\tiny {±0.3}} & 47.1 \scalebox{1.0}{\tiny {±1.9}} & 40.8 \scalebox{1.0}{\tiny {±0.6}} & 34.5 \scalebox{1.0}{\tiny {±0.8}} & 59.5 \scalebox{1.0}{\tiny {±0.7}} & 39.3 \scalebox{1.0}{\tiny {±0.6}} & 47.5 \\

\hline

\multicolumn{12}{c}{\textbf{Chinese}} \\
BioMistral 7B & \textbf{38.9} \scalebox{1.0}{\tiny {±5.5}} & 32.2 \scalebox{1.0}{\tiny {±5.5}} & 30.6 \scalebox{1.0}{\tiny {±2.2}} & \underline{31.9} \scalebox{1.0}{\tiny {±2.1}} & 30.1 \scalebox{1.0}{\tiny {±5.4}} & 29.3 \scalebox{1.0}{\tiny {±3.2}} & 27.8 \scalebox{1.0}{\tiny {±1.6}} & 22.8 \scalebox{1.0}{\tiny {±2.4}} & 57.5 \scalebox{1.0}{\tiny {±3.0}} & 29.7 \scalebox{1.0}{\tiny {±2.6}} & 33.1 \\
Mistral 7B Instruct & 37.0 \scalebox{1.0}{\tiny {±4.7}} & 34.3 \scalebox{1.0}{\tiny {±3.3}} & \underline{30.7} \scalebox{1.0}{\tiny {±3.9}} & 27.7 \scalebox{1.0}{\tiny {±3.1}} & 30.8 \scalebox{1.0}{\tiny {±5.4}} & \underline{29.9} \scalebox{1.0}{\tiny {±3.1}} & 28.5 \scalebox{1.0}{\tiny {±2.3}} & 23.4 \scalebox{1.0}{\tiny {±1.6}} & 58.1 \scalebox{1.0}{\tiny {±4.6}} & \underline{31.5} \scalebox{1.0}{\tiny {±1.5}} & 33.2 \\
\hdashline
BioMistral 7B DARE & \underline{38.6} \scalebox{1.0}{\tiny {±5.0}} & \underline{35.3} \scalebox{1.0}{\tiny {±6.3}} & 29.8 \scalebox{1.0}{\tiny {±2.5}} & 26.8 \scalebox{1.0}{\tiny {±2.8}} & \textbf{32.3} \scalebox{1.0}{\tiny {±7.2}} & 28.2 \scalebox{1.0}{\tiny {±5.4}} & \textbf{29.3} \scalebox{1.0}{\tiny {±2.2}} & \textbf{24.3} \scalebox{1.0}{\tiny {±2.7}} & 59.2 \scalebox{1.0}{\tiny {±5.1}} & \textbf{31.6} \scalebox{1.0}{\tiny {±2.2}} & \underline{33.6} \\
BioMistral 7B TIES & \underline{38.6} \scalebox{1.0}{\tiny {±5.6}} & 32.7 \scalebox{1.0}{\tiny {±5.1}} & \underline{30.7} \scalebox{1.0}{\tiny {±1.3}} & 30.1 \scalebox{1.0}{\tiny {±1.7}} & 30.3 \scalebox{1.0}{\tiny {±6.5}} & 28.8 \scalebox{1.0}{\tiny {±1.5}} & 28.4 \scalebox{1.0}{\tiny {±1.8}} & 24.0 \scalebox{1.0}{\tiny {±2.0}} & \underline{59.4} \scalebox{1.0}{\tiny {±4.5}} & 30.1 \scalebox{1.0}{\tiny {±2.6}} & 33.3 \\
BioMistral 7B SLERP & 37.5 \scalebox{1.0}{\tiny {±5.8}} & \textbf{35.5} \scalebox{1.0}{\tiny {±4.3}} & \textbf{31.9} \scalebox{1.0}{\tiny {±4.5}} & 30.0 \scalebox{1.0}{\tiny {±2.3}} & \underline{31.1} \scalebox{1.0}{\tiny {±7.6}} & \textbf{30.0} \scalebox{1.0}{\tiny {±5.9}} & \underline{29.2} \scalebox{1.0}{\tiny {±1.9}} & \underline{24.1} \scalebox{1.0}{\tiny {±3.4}} & \textbf{60.0} \scalebox{1.0}{\tiny {±4.7}} & \underline{31.5} \scalebox{1.0}{\tiny {±2.0}} & \textbf{34.1} \\
\hdashline
MedAlpaca 7B & 29.2 \scalebox{1.0}{\tiny {±3.4}} & 30.2 \scalebox{1.0}{\tiny {±4.0}} & 29.8 \scalebox{1.0}{\tiny {±1.8}} & \textbf{33.7} \scalebox{1.0}{\tiny {±4.6}} & 25.1 \scalebox{1.0}{\tiny {±1.2}} & 24.5 \scalebox{1.0}{\tiny {±2.3}} & 25.0 \scalebox{1.0}{\tiny {±0.8}} & 21.4 \scalebox{1.0}{\tiny {±1.2}} & 31.4 \scalebox{1.0}{\tiny {±15.2}} & 27.2 \scalebox{1.0}{\tiny {±0.3}} & 27.7 \\
PMC-LLaMA 7B & 24.2 \scalebox{1.0}{\tiny {±1.3}} & 27.3 \scalebox{1.0}{\tiny {±3.9}} & 30.2 \scalebox{1.0}{\tiny {±3.9}} & 18.6 \scalebox{1.0}{\tiny {±1.1}} & 26.0 \scalebox{1.0}{\tiny {±2.7}} & 24.0 \scalebox{1.0}{\tiny {±1.1}} & 26.3 \scalebox{1.0}{\tiny {±0.9}} & 20.6 \scalebox{1.0}{\tiny {±0.7}} & 32.3 \scalebox{1.0}{\tiny {±16.8}} & 24.8 \scalebox{1.0}{\tiny {±0.7}} & 25.4 \\
MediTron-7B & 25.8 \scalebox{1.0}{\tiny {±1.2}} & 30.2 \scalebox{1.0}{\tiny {±3.2}} & 29.0 \scalebox{1.0}{\tiny {±1.4}} & 17.8 \scalebox{1.0}{\tiny {±3.0}} & 26.7 \scalebox{1.0}{\tiny {±1.9}} & 24.1 \scalebox{1.0}{\tiny {±2.6}} & 27.4 \scalebox{1.0}{\tiny {±0.9}} & 21.3 \scalebox{1.0}{\tiny {±1.0}} & 52.1 \scalebox{1.0}{\tiny {±1.0}} & 29.0 \scalebox{1.0}{\tiny {±0.7}} & 28.3 \\
BioMedGPT-LM-7B & 30.3 \scalebox{1.0}{\tiny {±5.2}} & 28.0 \scalebox{1.0}{\tiny {±2.9}} & 29.4 \scalebox{1.0}{\tiny {±3.1}} & 24.1 \scalebox{1.0}{\tiny {±1.9}} & 29.3 \scalebox{1.0}{\tiny {±2.7}} & 28.8 \scalebox{1.0}{\tiny {±1.7}} & 27.0 \scalebox{1.0}{\tiny {±1.0}} & 22.9 \scalebox{1.0}{\tiny {±1.3}} & 56.5 \scalebox{1.0}{\tiny {±1.6}} & 27.7 \scalebox{1.0}{\tiny {±0.4}} & 30.4 \\
\hdashline
GPT-3.5 Turbo 1106 & 55.2 \scalebox{1.0}{\tiny {±3.6}} & 44.0 \scalebox{1.0}{\tiny {±2.2}} & 47.2 \scalebox{1.0}{\tiny {±0.3}} & 47.2 \scalebox{1.0}{\tiny {±0.8}} & 48.4 \scalebox{1.0}{\tiny {±2.0}} & 43.4 \scalebox{1.0}{\tiny {±2.9}} & 40.0 \scalebox{1.0}{\tiny {±1.3}} & 32.2 \scalebox{1.0}{\tiny {±1.0}} & 58.9 \scalebox{1.0}{\tiny {±0.1}} & 35.5 \scalebox{1.0}{\tiny {±0.3}} & 45.2 \\

\hline

\multicolumn{12}{c}{\textbf{French}} \\
BioMistral 7B & 42.5 \scalebox{1.0}{\tiny {±6.9}} & 38.2 \scalebox{1.0}{\tiny {±9.7}} & \underline{35.6} \scalebox{1.0}{\tiny {±7.3}} & \underline{36.2} \scalebox{1.0}{\tiny {±6.2}} & 33.1 \scalebox{1.0}{\tiny {±6.1}} & \textbf{35.5} \scalebox{1.0}{\tiny {±9.2}} & 30.7 \scalebox{1.0}{\tiny {±4.4}} & 25.2 \scalebox{1.0}{\tiny {±3.9}} & 61.5 \scalebox{1.0}{\tiny {±6.1}} & 32.5 \scalebox{1.0}{\tiny {±4.5}} & 37.1 \\
Mistral 7B Instruct & 39.7 \scalebox{1.0}{\tiny {±5.4}} & 38.1 \scalebox{1.0}{\tiny {±6.1}} & \underline{35.6} \scalebox{1.0}{\tiny {±7.7}} & 32.5 \scalebox{1.0}{\tiny {±7.2}} & 32.7 \scalebox{1.0}{\tiny {±5.2}} & 33.8 \scalebox{1.0}{\tiny {±6.3}} & 30.4 \scalebox{1.0}{\tiny {±3.3}} & 25.2 \scalebox{1.0}{\tiny {±2.9}} & 62.0 \scalebox{1.0}{\tiny {±6.7}} & 33.5 \scalebox{1.0}{\tiny {±3.1}} & 36.3 \\
\hdashline
BioMistral 7B DARE & \textbf{42.9} \scalebox{1.0}{\tiny {±7.3}} & \underline{39.8} \scalebox{1.0}{\tiny {±8.1}} & 34.6 \scalebox{1.0}{\tiny {±7.1}} & 31.8 \scalebox{1.0}{\tiny {±7.4}} & \textbf{35.3} \scalebox{1.0}{\tiny {±7.2}} & 33.9 \scalebox{1.0}{\tiny {±9.2}} & \underline{31.8} \scalebox{1.0}{\tiny {±4.0}} & \underline{26.5} \scalebox{1.0}{\tiny {±3.8}} & \underline{63.8} \scalebox{1.0}{\tiny {±7.6}} & \underline{34.3} \scalebox{1.0}{\tiny {±4.1}} & \underline{37.5} \\
BioMistral 7B TIES & \textbf{42.9} \scalebox{1.0}{\tiny {±7.6}} & 37.9 \scalebox{1.0}{\tiny {±8.6}} & 35.3 \scalebox{1.0}{\tiny {±6.6}} & 33.9 \scalebox{1.0}{\tiny {±5.5}} & 32.9 \scalebox{1.0}{\tiny {±6.5}} & \underline{35.2} \scalebox{1.0}{\tiny {±9.1}} & 31.2 \scalebox{1.0}{\tiny {±4.3}} & 26.2 \scalebox{1.0}{\tiny {±3.5}} & 63.0 \scalebox{1.0}{\tiny {±6.3}} & 33.0 \scalebox{1.0}{\tiny {±4.7}} & 37.2 \\
BioMistral 7B SLERP & \underline{42.6} \scalebox{1.0}{\tiny {±8.7}} & \textbf{40.2} \scalebox{1.0}{\tiny {±7.6}} & \textbf{37.0} \scalebox{1.0}{\tiny {±8.1}} & 35.3 \scalebox{1.0}{\tiny {±7.7}} & \underline{34.6} \scalebox{1.0}{\tiny {±7.9}} & 34.7 \scalebox{1.0}{\tiny {±8.3}} & \textbf{32.1} \scalebox{1.0}{\tiny {±4.3}} & \textbf{26.6} \scalebox{1.0}{\tiny {±4.5}} & \textbf{64.2} \scalebox{1.0}{\tiny {±7.0}} & \textbf{34.4} \scalebox{1.0}{\tiny {±4.4}} & \textbf{38.2} \\
\hdashline
MedAlpaca 7B & 31.8 \scalebox{1.0}{\tiny {±4.7}} & 31.2 \scalebox{1.0}{\tiny {±3.9}} & 33.4 \scalebox{1.0}{\tiny {±5.5}} & \textbf{37.7} \scalebox{1.0}{\tiny {±6.8}} & 28.3 \scalebox{1.0}{\tiny {±4.6}} & 25.5 \scalebox{1.0}{\tiny {±2.5}} & 27.0 \scalebox{1.0}{\tiny {±3.1}} & 22.9 \scalebox{1.0}{\tiny {±2.3}} & 39.1 \scalebox{1.0}{\tiny {±16.5}} & 28.1 \scalebox{1.0}{\tiny {±1.3}} & 30.5 \\
PMC-LLaMA 7B & 23.4 \scalebox{1.0}{\tiny {±1.9}} & 25.8 \scalebox{1.0}{\tiny {±4.0}} & 30.9 \scalebox{1.0}{\tiny {±3.5}} & 18.0 \scalebox{1.0}{\tiny {±1.4}} & 26.7 \scalebox{1.0}{\tiny {±2.6}} & 24.2 \scalebox{1.0}{\tiny {±1.0}} & 26.6 \scalebox{1.0}{\tiny {±0.9}} & 20.8 \scalebox{1.0}{\tiny {±0.6}} & 38.8 \scalebox{1.0}{\tiny {±16.5}} & 24.3 \scalebox{1.0}{\tiny {±0.9}} & 26.0 \\
MediTron-7B & 26.8 \scalebox{1.0}{\tiny {±1.9}} & 31.1 \scalebox{1.0}{\tiny {±3.3}} & 31.0 \scalebox{1.0}{\tiny {±3.3}} & 19.4 \scalebox{1.0}{\tiny {±3.4}} & 27.4 \scalebox{1.0}{\tiny {±1.9}} & 23.6 \scalebox{1.0}{\tiny {±2.4}} & 28.6 \scalebox{1.0}{\tiny {±1.9}} & 21.6 \scalebox{1.0}{\tiny {±1.0}} & 52.4 \scalebox{1.0}{\tiny {±1.0}} & 29.6 \scalebox{1.0}{\tiny {±1.0}} & 29.1 \\
BioMedGPT-LM-7B & 32.8 \scalebox{1.0}{\tiny {±5.6}} & 31.7 \scalebox{1.0}{\tiny {±5.9}} & 32.2 \scalebox{1.0}{\tiny {±4.7}} & 26.5 \scalebox{1.0}{\tiny {±3.8}} & 32.5 \scalebox{1.0}{\tiny {±5.4}} & 31.1 \scalebox{1.0}{\tiny {±3.6}} & 28.8 \scalebox{1.0}{\tiny {±2.7}} & 24.2 \scalebox{1.0}{\tiny {±2.2}} & 57.1 \scalebox{1.0}{\tiny {±1.6}} & 28.5 \scalebox{1.0}{\tiny {±1.2}} & 32.5 \\
\hdashline
GPT-3.5 Turbo 1106 & 63.4 \scalebox{1.0}{\tiny {±0.3}} & 65.3 \scalebox{1.0}{\tiny {±2.9}} & 58.8 \scalebox{1.0}{\tiny {±0.7}} & 63.4 \scalebox{1.0}{\tiny {±2.4}} & 59.0 \scalebox{1.0}{\tiny {±1.0}} & 54.5 \scalebox{1.0}{\tiny {±3.3}} & 49.0 \scalebox{1.0}{\tiny {±0.2}} & 42.3 \scalebox{1.0}{\tiny {±0.5}} & 63.3 \scalebox{1.0}{\tiny {±0.7}} & 46.2 \scalebox{1.0}{\tiny {±0.8}} & 56.5 \\

\hline

\multicolumn{12}{c}{\textbf{German}}  \\
BioMistral 7B & 45.1 \scalebox{1.0}{\tiny {±7.6}} & 39.5 \scalebox{1.0}{\tiny {±8.8}} & 36.8 \scalebox{1.0}{\tiny {±6.9}} & \underline{38.5} \scalebox{1.0}{\tiny {±6.7}} & 35.3 \scalebox{1.0}{\tiny {±6.5}} & \textbf{37.3} \scalebox{1.0}{\tiny {±8.6}} & 32.4 \scalebox{1.0}{\tiny {±4.8}} & 26.5 \scalebox{1.0}{\tiny {±4.1}} & 61.6 \scalebox{1.0}{\tiny {±5.3}} & 33.6 \scalebox{1.0}{\tiny {±4.3}} & 38.7 \\
Mistral 7B Instruct & 41.5 \scalebox{1.0}{\tiny {±5.7}} & 39.7 \scalebox{1.0}{\tiny {±6.0}} & 37.2 \scalebox{1.0}{\tiny {±7.2}} & 34.3 \scalebox{1.0}{\tiny {±7.0}} & 34.4 \scalebox{1.0}{\tiny {±5.4}} & 34.4 \scalebox{1.0}{\tiny {±5.6}} & 31.6 \scalebox{1.0}{\tiny {±3.5}} & 26.0 \scalebox{1.0}{\tiny {±2.9}} & 63.2 \scalebox{1.0}{\tiny {±6.2}} & 34.3 \scalebox{1.0}{\tiny {±3.0}} & 37.6 \\
\hdashline
BioMistral 7B DARE & 45.1 \scalebox{1.0}{\tiny {±7.4}} & \textbf{42.5} \scalebox{1.0}{\tiny {±8.6}} & \underline{37.4} \scalebox{1.0}{\tiny {±7.9}} & 34.6 \scalebox{1.0}{\tiny {±8.1}} & \textbf{37.1} \scalebox{1.0}{\tiny {±7.0}} & 35.2 \scalebox{1.0}{\tiny {±8.2}} & \textbf{33.7} \scalebox{1.0}{\tiny {±4.7}} & \textbf{28.0} \scalebox{1.0}{\tiny {±4.2}} & \underline{64.4} \scalebox{1.0}{\tiny {±6.7}} & \underline{35.3} \scalebox{1.0}{\tiny {±4.0}} & \underline{39.3} \\
BioMistral 7B TIES & \underline{45.5} \scalebox{1.0}{\tiny {±8.2}} & 39.6 \scalebox{1.0}{\tiny {±8.1}} & 36.8 \scalebox{1.0}{\tiny {±6.3}} & 36.4 \scalebox{1.0}{\tiny {±6.5}} & 35.1 \scalebox{1.0}{\tiny {±6.9}} & \underline{36.6} \scalebox{1.0}{\tiny {±8.3}} & \underline{32.8} \scalebox{1.0}{\tiny {±4.6}} & 27.3 \scalebox{1.0}{\tiny {±3.6}} & 62.3 \scalebox{1.0}{\tiny {±5.6}} & 34.1 \scalebox{1.0}{\tiny {±4.5}} & 38.7 \\
BioMistral 7B SLERP & \textbf{45.8} \scalebox{1.0}{\tiny {±9.4}} & \underline{42.4} \scalebox{1.0}{\tiny {±7.6}} & \textbf{39.1} \scalebox{1.0}{\tiny {±8.0}} & 37.5 \scalebox{1.0}{\tiny {±7.7}} & \underline{36.6} \scalebox{1.0}{\tiny {±7.7}} & 36.3 \scalebox{1.0}{\tiny {±7.7}} & \textbf{33.7} \scalebox{1.0}{\tiny {±4.7}} & \underline{27.8} \scalebox{1.0}{\tiny {±4.5}} & \textbf{65.1} \scalebox{1.0}{\tiny {±6.3}} & \textbf{35.4} \scalebox{1.0}{\tiny {±4.2}} & \textbf{40.0} \\
\hdashline
MedAlpaca 7B & 33.2 \scalebox{1.0}{\tiny {±4.8}} & 32.4 \scalebox{1.0}{\tiny {±4.6}} & 34.4 \scalebox{1.0}{\tiny {±5.1}} & \textbf{39.6} \scalebox{1.0}{\tiny {±6.8}} & 31.0 \scalebox{1.0}{\tiny {±6.4}} & 27.8 \scalebox{1.0}{\tiny {±4.6}} & 27.6 \scalebox{1.0}{\tiny {±2.9}} & 23.4 \scalebox{1.0}{\tiny {±2.3}} & 42.5 \scalebox{1.0}{\tiny {±15.5}} & 28.4 \scalebox{1.0}{\tiny {±1.2}} & 32.0 \\
PMC-LLaMA 7B & 23.7 \scalebox{1.0}{\tiny {±1.9}} & 25.3 \scalebox{1.0}{\tiny {±3.7}} & 30.7 \scalebox{1.0}{\tiny {±3.9}} & 17.8 \scalebox{1.0}{\tiny {±1.5}} & 27.7 \scalebox{1.0}{\tiny {±2.9}} & 24.8 \scalebox{1.0}{\tiny {±1.4}} & 26.9 \scalebox{1.0}{\tiny {±1.0}} & 20.8 \scalebox{1.0}{\tiny {±0.7}} & 42.2 \scalebox{1.0}{\tiny {±15.5}} & 24.2 \scalebox{1.0}{\tiny {±0.8}} & 26.4 \\
MediTron-7B & 27.5 \scalebox{1.0}{\tiny {±2.2}} & 31.3 \scalebox{1.0}{\tiny {±3.0}} & 31.7 \scalebox{1.0}{\tiny {±3.3}} & 19.7 \scalebox{1.0}{\tiny {±3.0}} & 27.1 \scalebox{1.0}{\tiny {±1.9}} & 23.2 \scalebox{1.0}{\tiny {±2.3}} & 28.8 \scalebox{1.0}{\tiny {±1.7}} & 21.8 \scalebox{1.0}{\tiny {±1.0}} & 52.5 \scalebox{1.0}{\tiny {±0.9}} & 29.8 \scalebox{1.0}{\tiny {±1.0}} & 29.3 \\
BioMedGPT-LM-7B & 35.1 \scalebox{1.0}{\tiny {±6.3}} & 33.0 \scalebox{1.0}{\tiny {±5.6}} & 34.1 \scalebox{1.0}{\tiny {±5.4}} & 28.8 \scalebox{1.0}{\tiny {±5.2}} & 33.3 \scalebox{1.0}{\tiny {±5.0}} & 31.8 \scalebox{1.0}{\tiny {±3.4}} & 29.4 \scalebox{1.0}{\tiny {±2.6}} & 24.7 \scalebox{1.0}{\tiny {±2.1}} & 57.4 \scalebox{1.0}{\tiny {±1.5}} & 28.8 \scalebox{1.0}{\tiny {±1.1}} & 33.6 \\
\hdashline
GPT-3.5 Turbo 1106 & 59.9 \scalebox{1.0}{\tiny {±1.6}} & 54.7 \scalebox{1.0}{\tiny {±2.4}} & 50.9 \scalebox{1.0}{\tiny {±0.3}} & 56.3 \scalebox{1.0}{\tiny {±0.8}} & 54.6 \scalebox{1.0}{\tiny {±1.0}} & 47.5 \scalebox{1.0}{\tiny {±2.1}} & 45.2 \scalebox{1.0}{\tiny {±0.7}} & 38.2 \scalebox{1.0}{\tiny {±0.6}} & 60.4 \scalebox{1.0}{\tiny {±0.3}} & 40.8 \scalebox{1.0}{\tiny {±0.2}} & 50.8 \\

\hline

\multicolumn{12}{c}{\textbf{Portuguese}}   \\
BioMistral 7B & 44.9 \scalebox{1.0}{\tiny {±6.8}} & \underline{41.3} \scalebox{1.0}{\tiny {±8.7}} & 37.2 \scalebox{1.0}{\tiny {±6.2}} & \underline{40.1} \scalebox{1.0}{\tiny {±6.9}} & 35.7 \scalebox{1.0}{\tiny {±5.9}} & \textbf{38.2} \scalebox{1.0}{\tiny {±7.9}} & 33.3 \scalebox{1.0}{\tiny {±4.6}} & 27.2 \scalebox{1.0}{\tiny {±3.9}} & 62.3 \scalebox{1.0}{\tiny {±4.9}} & 34.2 \scalebox{1.0}{\tiny {±4.1}} & 39.4 \\
Mistral 7B Instruct & 42.2 \scalebox{1.0}{\tiny {±5.3}} & 40.9 \scalebox{1.0}{\tiny {±5.9}} & 37.7 \scalebox{1.0}{\tiny {±6.7}} & 35.4 \scalebox{1.0}{\tiny {±6.7}} & 34.4 \scalebox{1.0}{\tiny {±4.9}} & 35.6 \scalebox{1.0}{\tiny {±5.7}} & 31.9 \scalebox{1.0}{\tiny {±3.2}} & 26.5 \scalebox{1.0}{\tiny {±2.8}} & 64.1 \scalebox{1.0}{\tiny {±5.9}} & 34.7 \scalebox{1.0}{\tiny {±2.8}} & 38.3 \\
\hdashline
BioMistral 7B DARE & \underline{45.2} \scalebox{1.0}{\tiny {±6.6}} & \textbf{43.1} \scalebox{1.0}{\tiny {±7.9}} & \underline{38.0} \scalebox{1.0}{\tiny {±7.2}} & 36.4 \scalebox{1.0}{\tiny {±8.0}} & \textbf{37.7} \scalebox{1.0}{\tiny {±6.4}} & 36.9 \scalebox{1.0}{\tiny {±8.1}} & \underline{34.3} \scalebox{1.0}{\tiny {±4.4}} & \textbf{28.6} \scalebox{1.0}{\tiny {±4.0}} & \underline{65.6} \scalebox{1.0}{\tiny {±6.5}} & \underline{35.7} \scalebox{1.0}{\tiny {±3.7}} & \underline{40.1} \\
BioMistral 7B TIES & \underline{45.2} \scalebox{1.0}{\tiny {±7.4}} & \underline{41.3} \scalebox{1.0}{\tiny {±8.0}} & 37.5 \scalebox{1.0}{\tiny {±5.9}} & 38.2 \scalebox{1.0}{\tiny {±6.8}} & 35.2 \scalebox{1.0}{\tiny {±6.2}} & 37.3 \scalebox{1.0}{\tiny {±7.6}} & 33.8 \scalebox{1.0}{\tiny {±4.6}} & 27.9 \scalebox{1.0}{\tiny {±3.5}} & 63.3 \scalebox{1.0}{\tiny {±5.4}} & 34.6 \scalebox{1.0}{\tiny {±4.1}} & 39.4 \\
BioMistral 7B SLERP & \textbf{46.6} \scalebox{1.0}{\tiny {±8.6}} & \textbf{43.1} \scalebox{1.0}{\tiny {±7.0}} & \textbf{39.4} \scalebox{1.0}{\tiny {±7.2}} & 39.5 \scalebox{1.0}{\tiny {±8.0}} & \underline{37.5} \scalebox{1.0}{\tiny {±7.2}} & \underline{38.1} \scalebox{1.0}{\tiny {±7.8}} & \textbf{34.4} \scalebox{1.0}{\tiny {±4.4}} & \underline{28.4} \scalebox{1.0}{\tiny {±4.2}} & \textbf{66.1} \scalebox{1.0}{\tiny {±5.9}} & \textbf{36.0} \scalebox{1.0}{\tiny {±4.0}} & \textbf{40.9} \\
\hdashline
MedAlpaca 7B & 33.8 \scalebox{1.0}{\tiny {±4.5}} & 32.7 \scalebox{1.0}{\tiny {±4.3}} & 35.1 \scalebox{1.0}{\tiny {±4.8}} & \textbf{40.6} \scalebox{1.0}{\tiny {±6.4}} & 30.9 \scalebox{1.0}{\tiny {±5.7}} & 29.1 \scalebox{1.0}{\tiny {±5.0}} & 28.0 \scalebox{1.0}{\tiny {±2.7}} & 24.0 \scalebox{1.0}{\tiny {±2.5}} & 45.0 \scalebox{1.0}{\tiny {±14.7}} & 28.6 \scalebox{1.0}{\tiny {±1.1}}  & 32.8  \\
PMC-LLaMA 7B & 23.9 \scalebox{1.0}{\tiny {±1.7}} & 25.2 \scalebox{1.0}{\tiny {±3.4}} & 30.3 \scalebox{1.0}{\tiny {±3.7}} & 17.7 \scalebox{1.0}{\tiny {±1.8}} & 28.0 \scalebox{1.0}{\tiny {±2.7}} & 24.7 \scalebox{1.0}{\tiny {±1.5}} & 26.9 \scalebox{1.0}{\tiny {±0.9}} & 20.9 \scalebox{1.0}{\tiny {±0.8}} & 44.2 \scalebox{1.0}{\tiny {±14.4}} & 24.1 \scalebox{1.0}{\tiny {±0.8}} & 26.6 \\
MediTron-7B & 27.8 \scalebox{1.0}{\tiny {±2.1}} & 31.7 \scalebox{1.0}{\tiny {±2.9}} & 31.4 \scalebox{1.0}{\tiny {±3.1}} & 20.4 \scalebox{1.0}{\tiny {±3.1}} & 27.7 \scalebox{1.0}{\tiny {±2.2}} & 23.0 \scalebox{1.0}{\tiny {±2.1}} & 29.0 \scalebox{1.0}{\tiny {±1.6}} & 21.8 \scalebox{1.0}{\tiny {±1.0}} & 52.7 \scalebox{1.0}{\tiny {±0.9}} & 30.0 \scalebox{1.0}{\tiny {±1.0}} & 29.6 \\
BioMedGPT-LM-7B & 35.1 \scalebox{1.0}{\tiny {±5.6}} & 33.3 \scalebox{1.0}{\tiny {±5.1}} & 34.8 \scalebox{1.0}{\tiny {±5.0}} & 30.0 \scalebox{1.0}{\tiny {±5.2}} & 33.6 \scalebox{1.0}{\tiny {±4.6}} & 32.2 \scalebox{1.0}{\tiny {±3.3}} & 29.8 \scalebox{1.0}{\tiny {±2.5}} & 24.8 \scalebox{1.0}{\tiny {±1.9}} & 58.0 \scalebox{1.0}{\tiny {±1.8}} & 28.7 \scalebox{1.0}{\tiny {±1.0}} & 34.0 \\
\hdashline
GPT-3.5 Turbo 1106 & 60.8 \scalebox{1.0}{\tiny {±1.5}} & 60.8 \scalebox{1.0}{\tiny {±1.5}} & 53.8 \scalebox{1.0}{\tiny {±2.4}} & 58.1 \scalebox{1.0}{\tiny {±1.4}} & 56.2 \scalebox{1.0}{\tiny {±0.8}} & 57.3 \scalebox{1.0}{\tiny {±1.8}} & 45.6 \scalebox{1.0}{\tiny {±0.4}} & 39.1 \scalebox{1.0}{\tiny {±0.9}} & 61.5 \scalebox{1.0}{\tiny {±0.5}} & 43.6 \scalebox{1.0}{\tiny {±0.3}} & 53.7 \\

\hline

\multicolumn{12}{c}{\textbf{Russian}} \\
BioMistral 7B & 45.5 \scalebox{1.0}{\tiny {±6.4}} & 42.4 \scalebox{1.0}{\tiny {±8.3}} & 37.8 \scalebox{1.0}{\tiny {±5.9}} & \underline{39.1} \scalebox{1.0}{\tiny {±6.7}} & 37.2 \scalebox{1.0}{\tiny {±6.4}} & \textbf{39.0} \scalebox{1.0}{\tiny {±7.4}} & 33.1 \scalebox{1.0}{\tiny {±4.3}} & 27.0 \scalebox{1.0}{\tiny {±3.6}} & 62.9 \scalebox{1.0}{\tiny {±4.7}} & 34.2 \scalebox{1.0}{\tiny {±3.7}} & 39.8 \\
Mistral 7B Instruct & 43.0 \scalebox{1.0}{\tiny {±5.1}} & 40.9 \scalebox{1.0}{\tiny {±5.5}} & 38.3 \scalebox{1.0}{\tiny {±6.2}} & 34.8 \scalebox{1.0}{\tiny {±6.3}} & 34.9 \scalebox{1.0}{\tiny {±4.6}} & 36.1 \scalebox{1.0}{\tiny {±5.3}} & 32.0 \scalebox{1.0}{\tiny {±2.9}} & 26.4 \scalebox{1.0}{\tiny {±2.5}} & 63.9 \scalebox{1.0}{\tiny {±5.4}} & 34.6 \scalebox{1.0}{\tiny {±2.6}} & 38.5 \\
\hdashline
BioMistral 7B DARE & 45.7 \scalebox{1.0}{\tiny {±6.1}} & \underline{43.7} \scalebox{1.0}{\tiny {±7.3}} & \underline{38.4} \scalebox{1.0}{\tiny {±6.7}} & 35.7 \scalebox{1.0}{\tiny {±7.5}} & \textbf{39.2} \scalebox{1.0}{\tiny {±6.8}} & 37.7 \scalebox{1.0}{\tiny {±7.6}} & \underline{34.1} \scalebox{1.0}{\tiny {±4.1}} & \textbf{28.4} \scalebox{1.0}{\tiny {±3.6}} & \underline{65.8} \scalebox{1.0}{\tiny {±6.0}} & \underline{35.8} \scalebox{1.0}{\tiny {±3.4}} & \underline{40.5} \\
BioMistral 7B TIES & \underline{46.0} \scalebox{1.0}{\tiny {±7.0}} & 42.3 \scalebox{1.0}{\tiny {±7.7}} & 38.2 \scalebox{1.0}{\tiny {±5.7}} & 37.2 \scalebox{1.0}{\tiny {±6.6}} & 36.8 \scalebox{1.0}{\tiny {±6.7}} & 38.4 \scalebox{1.0}{\tiny {±7.4}} & 33.5 \scalebox{1.0}{\tiny {±4.2}} & 27.7 \scalebox{1.0}{\tiny {±3.2}} & 64.0 \scalebox{1.0}{\tiny {±5.2}} & 34.6 \scalebox{1.0}{\tiny {±3.8}} & 39.9 \\
BioMistral 7B SLERP & \textbf{47.0} \scalebox{1.0}{\tiny {±7.9}} & \textbf{44.3} \scalebox{1.0}{\tiny {±6.9}} & \textbf{39.5} \scalebox{1.0}{\tiny {±6.6}} & 38.6 \scalebox{1.0}{\tiny {±7.6}} & \underline{38.6} \scalebox{1.0}{\tiny {±7.0}} & \underline{38.9} \scalebox{1.0}{\tiny {±7.4}} & \textbf{34.3} \scalebox{1.0}{\tiny {±4.1}} & \underline{28.2} \scalebox{1.0}{\tiny {±3.9}} & \textbf{66.0} \scalebox{1.0}{\tiny {±5.4}} & \textbf{35.9} \scalebox{1.0}{\tiny {±3.6}} & \textbf{41.1} \\
\hdashline
MedAlpaca 7B & 34.3 \scalebox{1.0}{\tiny {±4.3}} & 32.2 \scalebox{1.0}{\tiny {±4.2}} & 35.0 \scalebox{1.0}{\tiny {±4.4}} & \textbf{40.7} \scalebox{1.0}{\tiny {±5.9}} & 30.4 \scalebox{1.0}{\tiny {±5.4}} & 29.2 \scalebox{1.0}{\tiny {±4.6}} & 27.7 \scalebox{1.0}{\tiny {±2.5}} & 23.8 \scalebox{1.0}{\tiny {±2.3}} & 46.1 \scalebox{1.0}{\tiny {±13.7}} & 28.4 \scalebox{1.0}{\tiny {±1.2}} & 32.8  \\
PMC-LLaMA 7B & 23.9 \scalebox{1.0}{\tiny {±1.6}} & 24.8 \scalebox{1.0}{\tiny {±3.3}} & 30.7 \scalebox{1.0}{\tiny {±3.5}} & 17.7 \scalebox{1.0}{\tiny {±1.8}} & 27.8 \scalebox{1.0}{\tiny {±2.6}} & 24.9 \scalebox{1.0}{\tiny {±1.4}} & 27.0 \scalebox{1.0}{\tiny {±0.9}} & 20.9 \scalebox{1.0}{\tiny {±0.8}} & 45.2 \scalebox{1.0}{\tiny {±13.3}} & 23.9 \scalebox{1.0}{\tiny {±0.8}} & 26.7 \\
MediTron-7B & 28.0 \scalebox{1.0}{\tiny {±2.0}} & 31.9 \scalebox{1.0}{\tiny {±3.0}} & 31.6 \scalebox{1.0}{\tiny {±3.1}} & 20.1 \scalebox{1.0}{\tiny {±2.9}} & 27.3 \scalebox{1.0}{\tiny {±2.3}} & 23.1 \scalebox{1.0}{\tiny {±2.0}} & 29.1 \scalebox{1.0}{\tiny {±1.6}} & 21.5 \scalebox{1.0}{\tiny {±1.1}} & 52.8 \scalebox{1.0}{\tiny {±0.9}} & 29.7 \scalebox{1.0}{\tiny {±1.1}} & 29.5 \\
BioMedGPT-LM-7B & 35.3 \scalebox{1.0}{\tiny {±5.2}} & 34.5 \scalebox{1.0}{\tiny {±5.7}} & 34.7 \scalebox{1.0}{\tiny {±4.7}} & 30.4 \scalebox{1.0}{\tiny {±4.9}} & 34.1 \scalebox{1.0}{\tiny {±4.5}} & 32.4 \scalebox{1.0}{\tiny {±3.0}} & 29.7 \scalebox{1.0}{\tiny {±2.3}} & 24.7 \scalebox{1.0}{\tiny {±1.8}} & 57.7 \scalebox{1.0}{\tiny {±1.8}} & 28.6 \scalebox{1.0}{\tiny {±1.0}} & 34.2 \\
\hdashline
GPT-3.5 Turbo 1106 & 56.9 \scalebox{1.0}{\tiny {±0.9}} & 53.3 \scalebox{1.0}{\tiny {±2.9}} & 51.1 \scalebox{1.0}{\tiny {±3.1}} & 52.7 \scalebox{1.0}{\tiny {±2.4}} & 49.8 \scalebox{1.0}{\tiny {±1.2}} & 55.5 \scalebox{1.0}{\tiny {±2.4}} & 41.0 \scalebox{1.0}{\tiny {±0.7}} & 34.6 \scalebox{1.0}{\tiny {±0.7}} & 59.1 \scalebox{1.0}{\tiny {±0.9}} & 40.2 \scalebox{1.0}{\tiny {±0.4}} & 49.4 \\

\hline

\multicolumn{12}{c}{\textbf{Spanish}} \\
BioMistral 7B & 45.9 \scalebox{1.0}{\tiny {±6.0}} & 42.6 \scalebox{1.0}{\tiny {±7.7}} & 38.2 \scalebox{1.0}{\tiny {±5.6}} & \underline{40.2} \scalebox{1.0}{\tiny {±6.9}} & 37.7 \scalebox{1.0}{\tiny {±6.0}} & \underline{39.5} \scalebox{1.0}{\tiny {±7.0}} & 33.7 \scalebox{1.0}{\tiny {±4.2}} & 27.4 \scalebox{1.0}{\tiny {±3.5}} & 63.7 \scalebox{1.0}{\tiny {±4.8}} & 34.6 \scalebox{1.0}{\tiny {±3.6}} & 40.4 \\
Mistral 7B Instruct & 43.6 \scalebox{1.0}{\tiny {±5.0}} & 41.5 \scalebox{1.0}{\tiny {±5.3}} & 39.0 \scalebox{1.0}{\tiny {±6.0}} & 36.2 \scalebox{1.0}{\tiny {±6.8}} & 35.8 \scalebox{1.0}{\tiny {±4.9}} & 36.4 \scalebox{1.0}{\tiny {±5.0}} & 32.3 \scalebox{1.0}{\tiny {±2.8}} & 26.6 \scalebox{1.0}{\tiny {±2.4}} & 64.7 \scalebox{1.0}{\tiny {±5.4}} & 35.0 \scalebox{1.0}{\tiny {±2.6}}  & 39.1 \\
\hdashline
BioMistral 7B DARE & 46.2 \scalebox{1.0}{\tiny {±5.9}} & \textbf{44.6} \scalebox{1.0}{\tiny {±7.1}} & \underline{39.4} \scalebox{1.0}{\tiny {±6.7}} & 37.3 \scalebox{1.0}{\tiny {±8.0}} & \textbf{40.0} \scalebox{1.0}{\tiny {±6.7}} & 38.4 \scalebox{1.0}{\tiny {±7.3}} & \underline{34.5} \scalebox{1.0}{\tiny {±3.9}} & \textbf{28.7} \scalebox{1.0}{\tiny {±3.5}} & \textbf{66.8} \scalebox{1.0}{\tiny {±6.1}} & \underline{36.2} \scalebox{1.0}{\tiny {±3.2}} & \underline{41.2} \\
BioMistral 7B TIES & \underline{46.5} \scalebox{1.0}{\tiny {±6.5}} & 42.9 \scalebox{1.0}{\tiny {±7.3}} & 38.6 \scalebox{1.0}{\tiny {±5.3}} & 38.5 \scalebox{1.0}{\tiny {±7.0}} & 37.4 \scalebox{1.0}{\tiny {±6.3}} & 39.0 \scalebox{1.0}{\tiny {±7.0}} & 34.1 \scalebox{1.0}{\tiny {±4.1}} & 28.1 \scalebox{1.0}{\tiny {±3.1}} & \underline{64.8} \scalebox{1.0}{\tiny {±5.2}} & 35.1 \scalebox{1.0}{\tiny {±3.7}} & 40.5 \\
BioMistral 7B SLERP & \textbf{47.5} \scalebox{1.0}{\tiny {±7.5}} & \underline{44.5} \scalebox{1.0}{\tiny {±6.5}} & \textbf{39.9} \scalebox{1.0}{\tiny {±6.2}} & 39.8 \scalebox{1.0}{\tiny {±7.6}} & \underline{39.6} \scalebox{1.0}{\tiny {±7.0}} & \textbf{39.6} \scalebox{1.0}{\tiny {±7.1}} & \textbf{34.6} \scalebox{1.0}{\tiny {±3.9}} & \underline{28.6} \scalebox{1.0}{\tiny {±3.7}} & \textbf{66.8} \scalebox{1.0}{\tiny {±5.4}} & \textbf{36.3} \scalebox{1.0}{\tiny {±3.6}} & \textbf{41.7} \\
\hdashline
MedAlpaca 7B & 34.8 \scalebox{1.0}{\tiny {±4.3}} & 31.9 \scalebox{1.0}{\tiny {±4.1}} & 35.6 \scalebox{1.0}{\tiny {±4.4}} & \textbf{41.5} \scalebox{1.0}{\tiny {±5.8}} & 30.4 \scalebox{1.0}{\tiny {±5.0}} & 30.1 \scalebox{1.0}{\tiny {±4.8}} & 28.1 \scalebox{1.0}{\tiny {±2.5}} & 24.0 \scalebox{1.0}{\tiny {±2.2}} & 47.4 \scalebox{1.0}{\tiny {±13.0}} & 28.5 \scalebox{1.0}{\tiny {±1.1}} & 33.2 \\
PMC-LLaMA 7B & 24.0 \scalebox{1.0}{\tiny {±1.7}} & 24.2 \scalebox{1.0}{\tiny {±3.4}} & 30.6 \scalebox{1.0}{\tiny {±3.3}} & 17.5 \scalebox{1.0}{\tiny {±1.8}} & 27.7 \scalebox{1.0}{\tiny {±2.5}} & 25.0 \scalebox{1.0}{\tiny {±1.5}} & 27.0 \scalebox{1.0}{\tiny {±0.9}} & 21.0 \scalebox{1.0}{\tiny {±0.8}} & 46.3 \scalebox{1.0}{\tiny {±12.6}} & 23.8 \scalebox{1.0}{\tiny {±0.8}} & 26.7 \\
MediTron-7B & 28.4 \scalebox{1.0}{\tiny {±2.2}} & 31.9 \scalebox{1.0}{\tiny {±2.9}} & 31.9 \scalebox{1.0}{\tiny {±3.0}} & 21.1 \scalebox{1.0}{\tiny {±3.6}} & 28.1 \scalebox{1.0}{\tiny {±3.0}} & 23.3 \scalebox{1.0}{\tiny {±1.9}} & 29.2 \scalebox{1.0}{\tiny {±1.6}} & 21.6 \scalebox{1.0}{\tiny {±1.1}} & 53.0 \scalebox{1.0}{\tiny {±1.0}} & 29.8 \scalebox{1.0}{\tiny {±1.1}} & 29.8 \\
BioMedGPT-LM-7B & 35.5 \scalebox{1.0}{\tiny {±4.9}} & 34.8 \scalebox{1.0}{\tiny {±5.5}} & 35.0 \scalebox{1.0}{\tiny {±4.4}} & 31.7 \scalebox{1.0}{\tiny {±5.6}} & 34.2 \scalebox{1.0}{\tiny {±4.2}} & 32.7 \scalebox{1.0}{\tiny {±3.0}} & 30.0 \scalebox{1.0}{\tiny {±2.3}} & 24.7 \scalebox{1.0}{\tiny {±1.8}} & 58.1 \scalebox{1.0}{\tiny {±2.0}} & 28.6 \scalebox{1.0}{\tiny {±1.0}} & 34.5 \\
\hdashline
GPT-3.5 Turbo 1106 & 58.6 \scalebox{1.0}{\tiny {±0.2}} & 57.0 \scalebox{1.0}{\tiny {±1.4}} & 52.9 \scalebox{1.0}{\tiny {±0.3}} & 53.6 \scalebox{1.0}{\tiny {±0.9}} & 52.8 \scalebox{1.0}{\tiny {±0.3}} & 50.0 \scalebox{1.0}{\tiny {±1.4}} & 43.8 \scalebox{1.0}{\tiny {±0.2}} & 37.5 \scalebox{1.0}{\tiny {±0.5}} & 60.6 \scalebox{1.0}{\tiny {±0.5}} & 41.9 \scalebox{1.0}{\tiny {±0.2}} & 50.9 \\

\hline

\end{tabular}%
}
\caption{Models results using few-shot training on evaluation tasks translated into multiple languages. Scores are expressed in terms of accuracy ($\uparrow$).}
\label{tab:multilingual-results}
\end{table*}

\end{document}